\documentclass{article}

\usepackage{arxiv}

\usepackage[utf8]{inputenc} 
\usepackage[T1]{fontenc}    
\usepackage{hyperref}       
\usepackage{url}            
\usepackage{booktabs}       
\usepackage{amsfonts}       
\usepackage{nicefrac}       
\usepackage{microtype}      
\usepackage{lipsum}		
\usepackage{graphicx}
\usepackage{natbib}
\usepackage{doi}
\usepackage{multirow}
\usepackage{color}
\newcommand{\A}{\mathbf{A}}

\newcommand{\X}{\mathbf{X}}
\newcommand{\Y}{\mathbf{Y}}

\newcommand{\I}{\mathbf{I}}
\newcommand{\J}{\mathbf{J}}
\newcommand{\T}{\mathbf{T}}
\newcommand{\G}{\mathbf{G}}
\newcommand{\methodnamefull}{ModAlity Calibration}
\newcommand{\methodname}{MAC}

\title{MAC: ModAlity Calibration for Object Detection}

\author{  { Yutian Lei} \\	
	\texttt{yutianle@andrew.cmu.edu} \\
	\And
	 { Jun Liu} \\	
	\texttt{junliu2@andrew.cmu.edu} \\
 \And
	 { Dong Huang} \\	
	\texttt{donghuang@.cmu.edu} \\
 Robotics Institute, Carnegie Mellon University,	Pittsburgh, PA 15213 \\
}

\renewcommand{\shorttitle}{MAC: ModAlity Calibration for Object Detection}

\hypersetup{
pdftitle={A template for the arxiv style},
pdfsubject={q-bio.NC, q-bio.QM},
pdfauthor={David S.~Hippocampus, Elias D.~Striatum},
pdfkeywords={First keyword, Second keyword, More},
}

\begin{document}
\maketitle

\begin{abstract}
	The flourishing success of Deep Neural Networks(DNNs) on RGB-input perception tasks has opened unbounded possibilities for non-RGB-input perception tasks, such as object detection from wireless signals, lidar scans, and infrared images. Compared to the matured development pipeline of RGB-input (source modality) models, developing non-RGB-input (target-modality) models from scratch poses excessive challenges in the modality-specific network design/training tricks and labor in the target-modality annotation. In this paper, we propose ModAlity Calibration (MAC), an efficient pipeline for calibrating target-modality inputs to the DNN object detection models developed on the RGB (source) modality. We compose a target-modality-input model by adding a small calibrator module ahead of a source-modality model and introduce MAC training techniques to impose dense supervision on the calibrator. By leveraging (1) prior knowledge synthesized from the source-modality model and (2) paired \{target, source\} data with \textbf{zero} manual annotations, our target-modality models reach comparable or better metrics than baseline models that require 100\% manual annotations. We demonstrate the effectiveness of MAC by composing the WiFi-input, Lidar-input, and Thermal-Infrared-input models upon the pre-trained RGB-input models respectively.
\end{abstract}


\section{Introduction}
\label{sec:intro}
\begin{figure}
  \includegraphics[width=\textwidth]{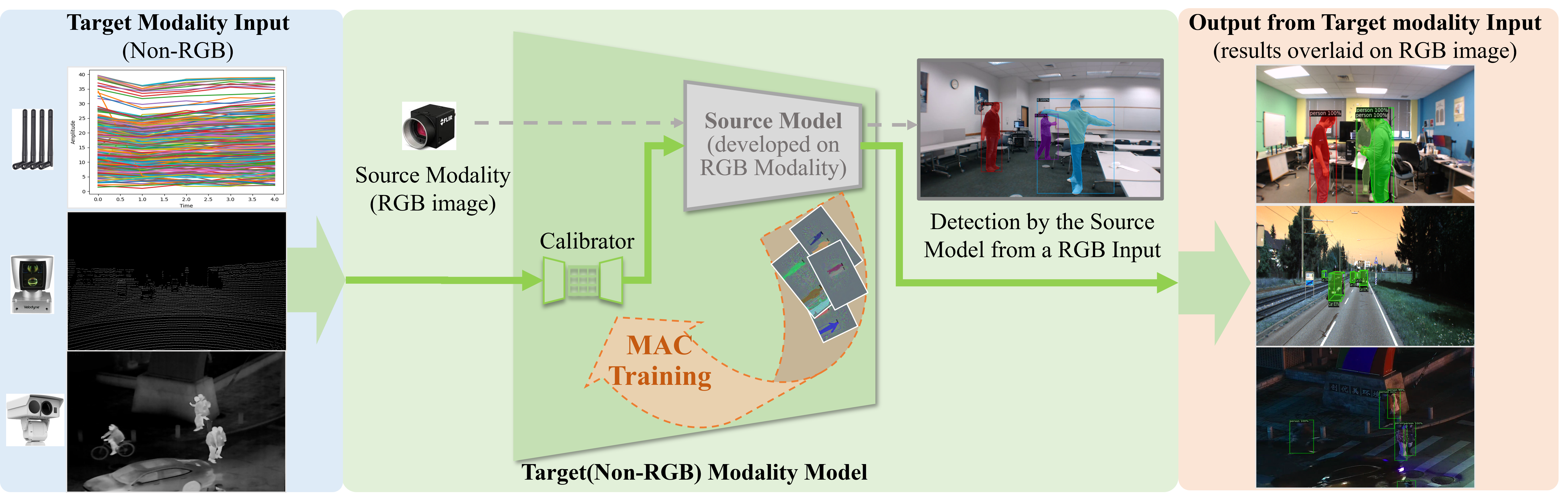}
  \caption{We propose \methodnamefull~(\methodname) for switching input modalities of DNN models. Our ``Target Modality Model" is constructed and trained upon a pre-trained source-modality (RGB-input) model by ``\methodname~Training". The dashed arrows ``\textbf{$\mathbf{\dashrightarrow}$}" are for training only. In inference, each target-modality(Non-RGB) input is processed along the thick solid green arrows ``\textcolor{green}{$\textbf{\textrightarrow}$}".}
  \label{fig:Teaser}
\end{figure}
Although research on the DNN-based perception problem has been largely focused on RGB-input models, there are wide scenarios in that non-RGB sensors have clear advantages over RGB cameras. For instance, wireless signals~\cite{zhao2018through,Wanghuang2019} can easily penetrate furniture occlusion and identify human bodies for their Dielectric properties, while being lighting-free, occlusion-resistant, and privacy-friendly compared to cameras. Lidar scans~\cite{Geiger2012CVPR,Sun2020Waymo} contain depth information, enabling more accurate and robust object localization than RGB under low light or bad weather. Thermal InfRared(TIR) cameras~\cite{hwang2015multispectral,jia2021llvip} capture near-infrared ($0.75$-$1.3\mu m$) or long-wavelength infrared ($7.5$-$13\mu m$) signals, which makes, in particular, human bodies more visible than in RGB images and more robust to the visible spectrum interference.

Thanks to the decade of work on RGB-input DNN models, researchers has accumulated extensive resources on image-appeal architectures(e.g., MaskRCNN~\cite{DBLP:journals/corr/HeGDG17}, Yolo5~\cite{glenn_jocher_2020_4154370}, Swin-Transformer~\cite{liu2021Swin}), pre-train-datasets~(ImageNet~\cite{imagenet_cvpr09}, MS-COCO~\cite{ms_coco}, OpenImages~\cite{OpenImages}), pre-train weights, training tricks~\cite{he2018bag,zhang2019bag,ghiasi2020simple} and code repos(e.g., Detectron2~\cite{wu2019detectron2}, mmDetection\cite{mmdetection}). Unfortunately, non-RGB-input models cannot be \textbf{directly} built upon above RGB resources. Instead, one usually needs new DNN designs\cite{qi2017pointnetplusplus,zhao2018through,Wanghuang2019}, from-scratch training, and new data collection/annotation of non-RGB sensors at the comparable scale of RGB databases above. 

In this paper, we propose \methodnamefull~(\methodname) for calibrating target-modality inputs to a DNN model developed on the source modality. Figure~\ref{fig:Teaser} shows the main idea of \methodname. Ahead of a source-modality model, we add a small target-modality-input calibrator module, composing a \textbf{[Calibrator$|$Source]}-structured target-modality model. The calibrator transforms a target-modality input into a source-modality-like tensor highlighting the foreground, which is then mapped to object detection results by the source module. Trained on \{source, target\} input pairs of \textbf{zero} manual annotation, the \methodname~target-modality models reach comparable or better metrics on WiFi, Lidar, and Thermal Infrared than the baselines that require 100\% manual annotation. This is achieved by our \methodname~training techniques that learn prior knowledge from the enclosed source module and iteratively regularize gradients on the calibrator layers. \methodname~training helps the detection task by mimicking the foreground features of the source modality inputs.

\textbf{Summary of Contributions:} 
\begin{itemize}
    \item MAC, A simple pipeline, is proposed for building Non-RGB-input models upon pre-trained RGB-input models in reducing the DNN design efforts and training data. 
    \item MAC training techniques are proposed to address the special vanishing gradient problem arising by adding a calibrator to a pre-trained RGB perception model. MAC training introduces strong and dense gradients that significantly improve the target-modality model metrics and reduce the need for annotations.
    \item Compared with target-modality-input models, (i.e., WiFi, Lidar, Thermal Infrared) under naive training, the MAC training techniques achieve comparable or better metrics without manual annotation (\methodname-self-supervised), and significantly better metrics with manual annotation(\methodname-supervised). 
\end{itemize}

\section{Realated Works} \label{chapBackground}
For conciseness, we only list object detection work on the three Non-RGB target modalities (WiFi, Lidar, and Thermal) involved in our experiments. 

\textbf{Modality-specific Perception.}
 In most two-stage approaches, Non-RGB inputs are converted to RGB images before feeding to an RGB-input model. Researchers of \cite{KatoFukushima2021,KefayatiPourahmadi2020,Drob2021} only convert Wifi signals to low-resolution ($<160 \times 120$) RGB images by over-fitting a few antenna layouts. No codes or data are available. Points2Pix ~\cite{MilzSimon2019} translates Lidar points to RGB images with a conditional GAN. The infrared images were re-colored to RGB in ~\cite{Limmer2019,Rajendran2019} by image translation ~\cite{Karras2019stylegan2}. 

In single-stage approaches, the whole model is only designed and trained on a non-RGB modality. Due to the lack of spatial representation, the WiFi-input models are mostly focused on coarse-granularity tasks such as crowd counting ~\cite{Depatla2018,liu2019deepcount} or single-person activity recognition ~\cite{WangLiu2017,LiHe2019}.  ~\cite{Wanghuang2019} develop pioneer WiFi-specific DNNs for multi-person segmentation and pose estimation. Lidar-input models are specific to point-clouds representations, such as the Point View ~\cite{qi2016pointnet,qi2017pointnetplusplus,Shi_2019_CVPR}, the Bird’s Eye View ~\cite{yan2018second,lang2019pointpillars} and the Range View(~\cite{Meyer2019LaserNet,Fan_2021_ICCV}). The range view is popular for its low quantization error and computational costs. Thermal images are close to RGB spatially, enabling  ~\cite{hwang2015multispectral,jia2021llvip} to train RGB-input models on the TIR inputs.

Unlike existing two-stage approaches, we work on perception tasks that learn foreground representation instead of RGB appearances. By enclosing the pre-train source model in the target model, we simplify the design effort and reduce target-modality annotations of the single-stage approaches. 

\textbf{Domain Adaption.} By definition\footnote{\url{https://en.wikipedia.org/wiki/Domain_adaptation}}, domain adaptation mainly addresses the shift of data sampling distribution. Modality adaption, on the other hand, mainly addresses the change in the spatial, temporal, and physical nature of the inputs. Nevertheless, some works still consider modality adaption as a special case of domain adaption especially in the case of image-to-image translation \cite{Murez2018, dou2018unsupervised, Pizzati2020,Musto2020,Xie2020,gal2021stylegannada}. All these works aim to generate realistic image pixels in new domains. 

Our work applies to any non-image modalities such as WiFi signals. Instead of pursuing realistic pixels, we ONLY generate foreground-associated features contributing to the detection tasks, which aims to reduce the efforts in DNN design and data annotation. Moreover, for simplicity, this work is presented under the assumption that the source and target modality data are sampled from the same foreground/background distributions, such that MAC is only focused on reducing the discrepancy between modalities.


\textbf{Knowledge Distillation between Models.} Teacher-to-student Knowledge Distillation (KD) is a popular approach ~\cite{hinton2015distilling,dhar2019learning}. The student models mimic the teacher models in predictive probabilities ~\cite{hinton2015distilling,li2017learning}, intermediate features \cite{romero2014fitnets,wang2019distilling}, or attention maps \cite{zagoruyko2016paying,wang2017residual,liu2020continual,chen2017learning}. When the teacher and student models have different input modalities ~\cite{Gupta2016}, KD requires the same amount of annotated source-target data as those in the teacher training. All KD methods run the teacher and student in parallel during training. 

Unlike KD, our [Calibrator$|$Source] target model is initialized and supervised by the enclosed Source module, which requires neither an independent teacher inference dataflow nor fully annotated target-modality data. Ablation study shows that \methodname~clearly performs better.

\textbf{Adversarial Training.} To improve robustness on imbalanced datasets, many adversarial training strategies explicitly produce hard features/samples: auto-augmentation ~\cite{Zoph2020}, co-mixup ~\cite{kim2021comixup}, random erasing ~\cite{zhong2017random}, representation self-challenging ~\cite{HuangIAN2020,huangRSC2020}, reverse attention ~\cite{chen2020tip}. Regulators such as cross-layer consistency \cite{hou2019learning,wang2019sharpen} and self-distillation mechanism ~\cite{huang2020comprehensive} were also very effective. Generative Adversarial Networks(GAN) implicitly produce hard samples from a discriminator and are recently extended to the object detection tasks by ~\cite{rabbi2020small,LiuMuelly2019}. 

We improve target model robustness by synthesizing image-like foreground representation and regularising gradients of the enclosed source model. 

\textbf{Vector Quantized Representation.} VQ-VAE ~\cite{VQVAE2017,VQVAE2019} and VQ-GAN ~\cite{VQGAN2021} show that a quantized latent space provides a compact representation of natural images, language, and audio/video sequence while using a relatively small number of parameters, making them efficient to train and use.

We extend VQVAE to learn the foreground representation shared between modalities.

\section{ModAlity Calibration (MAC)} \label{MAC}
\begin{figure*}[ht]
\centering
\includegraphics[width=.99\linewidth]{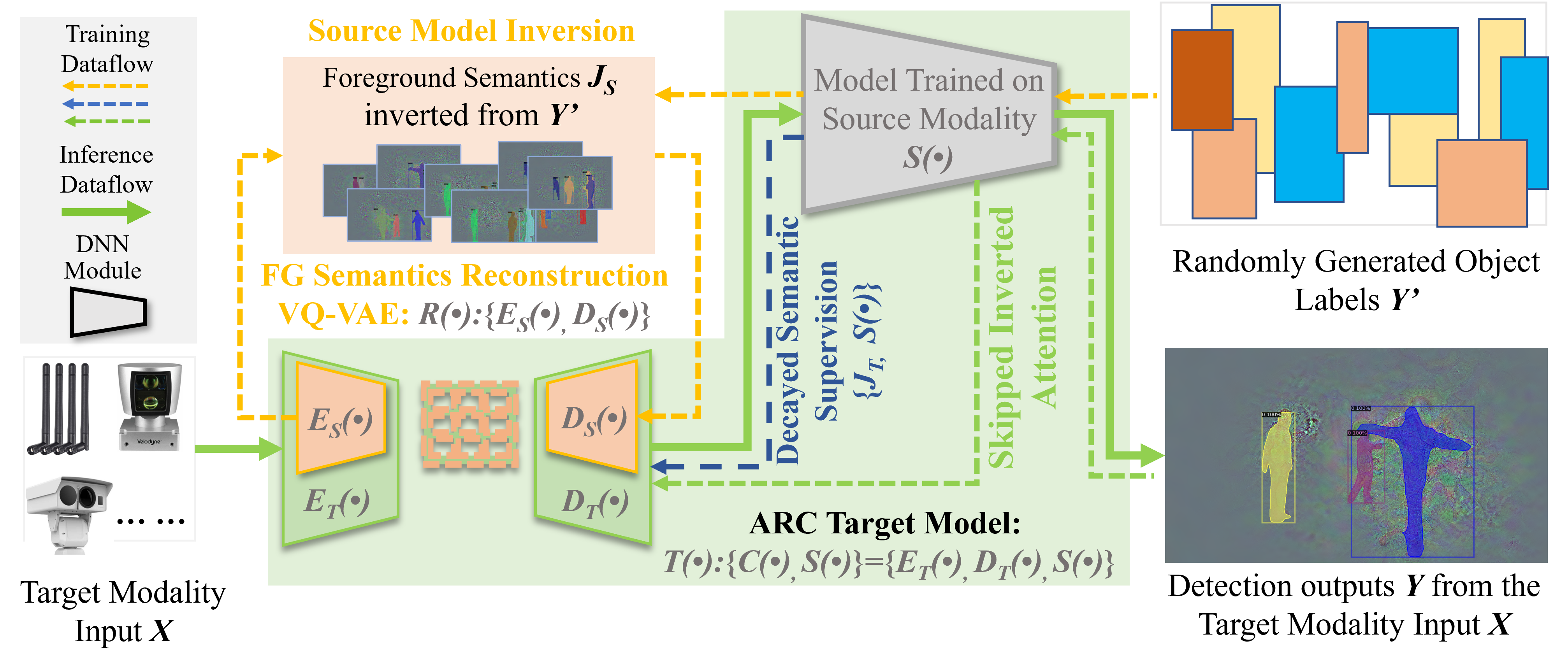}
\caption{\methodnamefull (\methodname) Framework. Our \methodname~target model $T(\cdot):\{E_{T}, D_{T}, S\}$ is composed of calibrator $C(\cdot):\{E_{T}, D_{T}\}$ and source model $S(\cdot)$, with its inference dataflow along the solid green arrows (``\textcolor{green}{$\mathbf{\longrightarrow}$}"). The \methodname~training includes three types of supervision(along the dashed arrows ``$\mathbf{\dashleftarrow}$") : \textbf{(a)} Foreground(FG) Semantic Reconstruction(\textbf{FSR}), which leverages the source model prior: the foreground(object) semantics $\J_{S}$ synthesized from $S(\cdot)$ and self-reconstructed by an auxiliary VQ-VAE, $R(\cdot):\{E_{S}, D_{T}\}$. \textbf{(b)} Decayed Semantic Supervision(\textbf{DSS}), which regularizes the supervision from precomputed foreground semantics $\J_T$ on target modality training data and from source model $S(\cdot)$. \textbf{(c)} Skipped Inverted Attention(\textbf{SIA}), which improves the attention of $C(\cdot)$ output using high-level $S(\cdot)$ layer gradients. See details in the \textbf{\methodname~Training} section.}
\label{fig:Framework}
\vspace{-0.05cm}
\end{figure*}
\textbf{Problem Definition}: \methodname~for the object detection task:
\begin{itemize}
    \item \textbf{Source model}: $S(\cdot): \I\rightarrow \Y$ maps one RGB image $\I \in \Re^{[width\times height\times 3]}$ to the object locations and categories $\Y= [object\_bboxes, object\_mask, object\_class]$. 
    \item \textbf{Target model}: $T(\cdot):  \X\rightarrow \Y$ maps one target modality tensor $\X\in \Re^{[width\_T\times height\_T\times channel\_T]}$ to $\Y$.
    \item \textbf{\methodname~target model}: $T(\cdot):\{C(\cdot)| S(\cdot)\}$, where the ``Calibrator" module $C(\cdot): \X\rightarrow \J$ produces an image-like tensor $\J\in \Re^{[width\times height\times 3]}$. The ``Source" module $S(\cdot)$ maps $\J$ to $\Y$.
\end{itemize}

The goal of \methodname~is to train a \methodname~target model $\{C(\cdot)|S(\cdot)\}$ given a pre-trained source model $S(\cdot)$ and a set of $\{\X, \I\}$ pairs.  (See the framework in Figure~\ref{fig:Framework}). 

For simplicity, we assume that the source and target modality data are sampled from the same $\Y$ (foreground/background) distributions, such that \textbf{\methodname ~is \textit{only} focused on reducing the discrepancy between modalities. }

\subsection{Reasoning for the $\{C(\cdot)|S(\cdot)\}$ Target Model}

If we follow the development procedure of $S(\cdot)$ to develop a new $T(\cdot)$, we need a modality-specific multi-resolution feature extractor (comparable to ResNet), which is coupled with a task-specific output head (Such as the anchor-based or transformer-based bounding box regressors) by multi-resolution skip connections. One also needs annotated $\{\X, \Y\}$ data comparable to the amount of the $\{\I, \Y\}$ data used in the source model $S(\cdot)$ training. 

Under \methodname (Figure~\ref{fig:Framework}), we only design calibrator $C(\cdot)$ that produces a \textit{single-resolution} tensor $\J$ feeding to the source module $S(\cdot)$ enclosed in $T(\cdot)$. 

\textbf{Foreground encoding in $C(\cdot)$:} Since it is $S(\cdot)$'s expertise to locate and classify objects, $C(\cdot)$ only needs to pass to $S(\cdot)$ some foreground-sensitive features $\J$, i.e., the edges/textures highlighting all the object categories. To generate such foreground features, we revisit the insight of the image-wise Class Activation Maps ~\cite{zhou2016learning}: all pixels of each object category can be mapped to an element of a probability vector by Softmax activation, which highlights foreground pixels assuming category-wise features follow a multi-modal distribution. In order to preserve the internal spatial layout of objects, we encode pixel patches(object parts) under the multi-modal distribution. This is done by a encoder-decoder structure $C(\cdot):\{E_{T}, D_{T}\}$ with a quantized latent space inspired by VQ-VAE\cite{VQVAE2017} (see Figure~\ref{fig:Framework}). We set the latent space tensor size to $[width/8, height/8, channel]$, in which every $[1\times 1\times channel]$ vector is hard-coded to one of the multi-modal centers of local patches, denoted as codebook $\{B_{i} \in \Re^{1\times 1\times channel}\}|_{i=1,\dots, p}$. Mapping such a VQ-VAE-like latent space to image-like feature $\J$, the calibrator decoder $D_{T}$ simply takes the same structure as the standard VQ-VAE decoder. The calibrator encoders $E_{T}$ for target-modality inputs are similar to the standard VQ-VAE encoder with minor adjustments below.



\begin{figure*}[!htb]
\centering
\includegraphics[width=.99\linewidth]{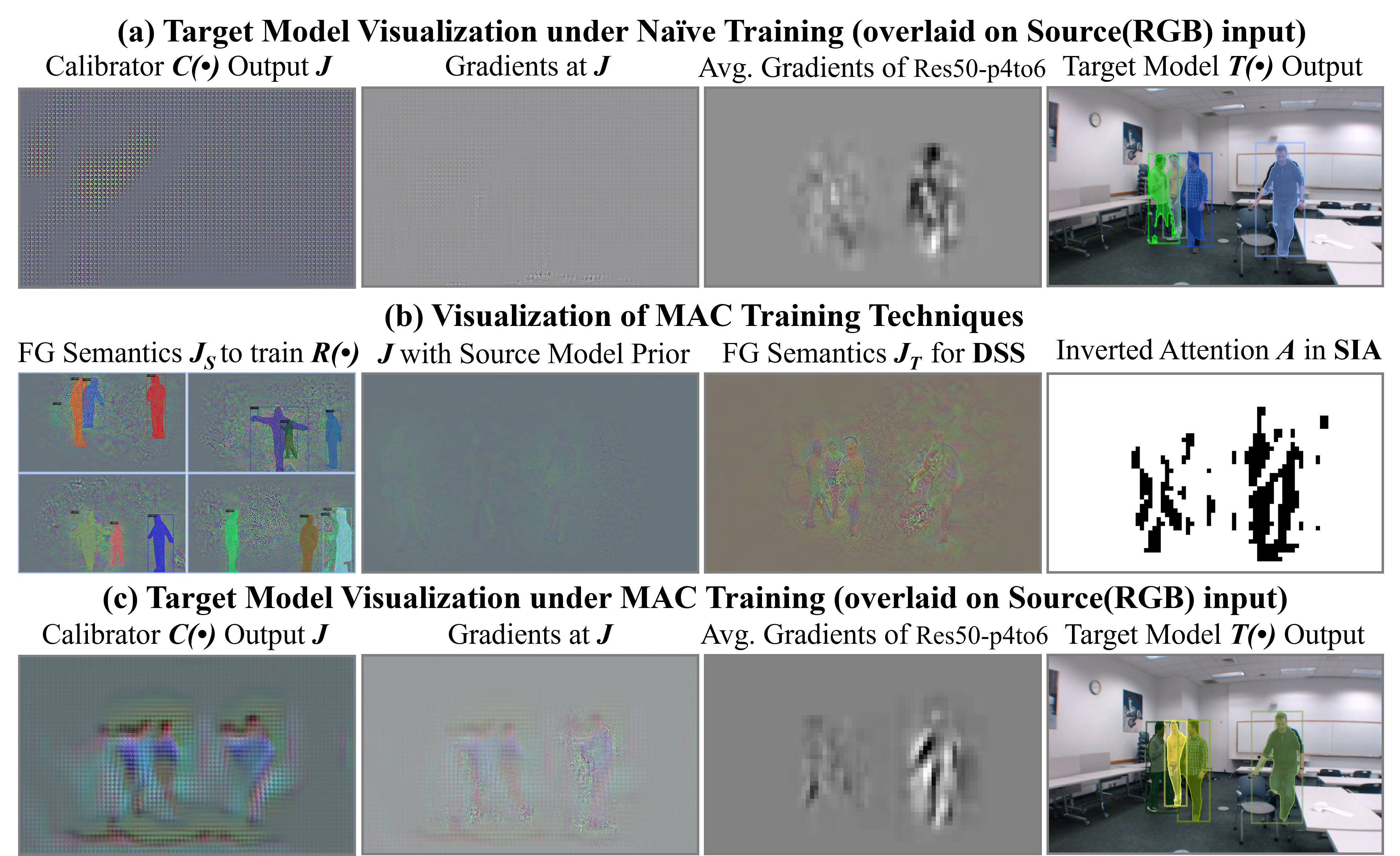}
\caption{Training strategy examples on a WiFi-input target model $T(\cdot):\{C(\cdot), S(\cdot)\}$. Here, the source model $S(\cdot)$ is an RGB-input ResNet50-FPN-Mask-RCNN. \textbf{(a)} Target model visualization under \textit{Na\"ive} training shows neither foreground-sensitive features in $C(\cdot)$ output $\J$ nor any clear gradients at $\J$ compared to ``Avg. Gradients of Res50-p4to6" (the high-level gradients visualized by resizing and averaging on one channel). \textbf{(b)} Visualization of MAC training techniques (described the \textbf{MAC Training} section). \textbf{(c)} Target model visualization under \methodname~training shows clear foreground-sensitive features, strong gradients at $\J$ and better detection. Note that, in ``Target Model $\T(\cdot)$ Output" of the 3rd row, all persons are detected with better masks than ``Target Model $\T(\cdot)$ Output" of the 1st row. Please zoom in to see all the instance masks. }
\label{fig:MAC_Train}
\end{figure*}

\textbf{Target-Modality Encoder $E_{T}$:} The WiFi signal corresponding to one synchronized RGB image ~\cite{Wanghuang2019}, is represented as the Channel State Information (CSI) ~\cite{halperin2011tool} tensor [samples, transmitters, receivers, sub-carriers]. The CSI tensor elements has no spatial dependence as RGB pixels. In this case, we construct $E_{T}$ by adding Wi2Vi ~\cite{KefayatiPourahmadi2020} layers in front of a VQ-VAE encoder. For other target modalities (infrared and Lidar range images) that have the image-like tensor shape, we directly use the VQ-VAE encoder structure for $E_{T}$. 

\textbf{Remarks:} To enable $T(\cdot):\{C(\cdot), S(\cdot)\}$ to produce the same $\Y$ as that of a pre-trained source model $S(\cdot)$, the latent space of $C(\cdot)$ has to contain the same semantics as those encoded in $S(\cdot)$. This opens the possibility to learn $C(\cdot)$ from $S(\cdot)$, without requiring a huge set of $\{\X, \Y\}$ training data. 

\subsection{MAC Training}

The devil lies in the training of the \methodname~target model $T(\cdot)$. There are two \textit{na\"ive} sources of supervision: \textbf{(1)} Given the $\{\X, \I\}$ pairs, one may pre-train $C(\cdot)$ to approximate $\I$. Since there are usually more background pixels than foreground pixels, the pre-trained $C(\cdot)$ outputs $\J$ may largely be background textures that are irrelevant to the detection tasks. \textbf{(2)} Given abundant $\{\X, \Y\}$ training pairs, one may randomly initialize $C(\cdot)$ and update all $T(\cdot)$ layers using the gradients back-propagated from the source model losses. Such a strategy suffers from the vanishing gradient problem ~\cite{Bengio1994,GlorotBengio2010} and is contradictory to the common DNN training practice (for instance, ImageNet-pretrained Resnet + randomly initialized Mask-RCNN). In fact, the randomly initialized $C(\cdot)$ should receive the strong gradients for updating, while the pre-trained weights of $S(\cdot)$ should be preserved by updating with weak gradients. Figure~\ref{fig:\methodname_Train}(a) shows that the \textit{na\"ive} training produces neither foreground-sensitive features nor any clear gradients at $\J$ comparable to the high-level gradients, e.g., ``Avg. Gradients of Res50-p4to6". 


To address the above issues, we propose the \methodname~training techniques below (Also see diagram in Figure~\ref{fig:Framework}).

\textbf{Road-map:} MAC training techniques address the special vanishing gradient problem arising by adding a calibrator to a pre-trained RGB perception model. Source Model Inversion(SMI) discovers the dense image-like foreground features. Foreground Semantics Reconstruction(FSR) introduces dense image-like supervision by pre-training the calibrator. Decayed Semantic Supervision (DSS) uses $\J_{T}$ to provide strong gradients for foreground supervision, then let the weaker gradients from $\mathcal{L}_{S}$ amend the acute details. Skipped Inverted Attention (SIA) channels the high-level gradient-based attention from $S(\cdot)$ back to the low-level feature layers of $C(\cdot)$. Along with section 3.2, Fig.~\ref{fig:MAC_Train} and Table~\ref{table:Ablation} demonstrate how the gradients are improved by each MAC training technique. 


\subsubsection{Source Model Inversion(SMI)}
We first guide the calibrator $C(\cdot)$ to produce foreground-sensitive features leveraging a pre-trained source model $S(\cdot)$. Given the pre-trained $S(\cdot)$ and the object annotation $\Y$, we conduct model inversion ~\cite{vondrick2013hoggles,CaoJohnson2021} to generate $\J_{S}$ by minimizing the $S(\cdot)$ losses, 
\begin{equation}
\mathcal{L}_{S}(S(\J_{S}), \Y)= \lambda_{bbox}\mathcal{L}_{bbox}+\lambda_{cls}\mathcal{L}_{cls}+\lambda_{mask}\mathcal{L}_{mask}.
\label{Eqn:L_Source}
\end{equation} 
This problem is solved by image-wise optimization: freezing all the $S(\cdot)$ layers, computing gradient from $\mathcal{L}_{S}(\J_{S}, \Y)$ and only updating $\J_{S}$ from its random initialization. After the optimization converges, $\J_{S}$ is synthesized as a most probable and style-invariant ``image" from which $S(\cdot)$ can detect $\Y$. $\J_{S}$ only captures the edge-like patterns of the foreground (See pixels marked in colors Figure~\ref{fig:MAC_Train}(b)). We call $\J_{S}$ the \textbf{Foreground Semantics}. 

Compared with the original RGB pixels, we see $\J_{S}$ a clean foreground-focused supervision to guide $C(\cdot)$ training. Compared with the small gradient amplitude from $\mathcal{L}_{S}$, the amplitude of $\J_{S}$ is comparable to $\I$ resulting in strong gradients on the $C(\cdot)$ layers. 

To increase the diversity of $\J_{S}$, we also randomly generate the object layouts in $\Y'$ of different bbox locations (instance masks for Mask-RCNN) and class labels. The random object labels $\Y$ introduce diverse object locations, sizes, and co-occurrences. In addition, the noise diversity of the foreground is also introduced by the random $\J$ initialization of model inversion from different $\Y'$s.

\subsubsection{Foreground Semantics Reconstruction(FSR)} 
 Next, to inject the prior knowledge in $J_{S}$ to calibrator $C(\cdot)$, we train an auxiliary VQ-VAE, $R(\cdot)=\{E_{S}, D_{S}\}: \J_{S} \rightarrow \J_{S}$ that encodes and reconstructs $J_{S}$. Then we share the $R(\cdot)$'s VQ codebook with $C(\cdot)$, and initialize the $C(\cdot)$'s decoder $D_{T}(\cdot)$ by the $R(\cdot)$'s decoder $D_{S}(\cdot)$ weights. We call the initialization method of $C(\cdot)$ as \textbf{Foreground Semantics Reconstruction(FSR)}. 

In addition, our target model $T(\cdot)$ encloses $S(\cdot)$ as a module, which can explicitly inherit the source model knowledge by initializing it with the pre-trained source model weights.

Finally, to accommodate both above priors incorporated into $C(\cdot)$ and $S(\cdot)$, we train $T(\cdot)$ in a \textbf{two-stage update} strategy: (i) fix $S(\cdot)$ and only update $C(\cdot)$ until it converges; (ii) continue training by updating both $C(\cdot)$ and $S(\cdot)$. 

\textbf{Remarks:} \textbf{(i)} $\J_{S}$ is synthesized from a pre-trained source model over synthetic $\Y'$. No manually annotated source inputs $\I$ are needed. \textbf{(ii)} Given that the $R(\cdot)$ training requires no annotation on $\X$, and the $S(\cdot)$ module can provide pseudo ground truth $\Y_{pseudo}$ for the $\{\X, \I\}$ pairs, the overall training of $T(\cdot)$ is \textbf{self-supervised} (\textbf{zero} manual annotation on $\X$). 

\subsubsection{Decayed Semantic Supervision (DSS)} 
The Decayed Semantic Supervision is used to regularize $\mathcal{L}_{S}$ gradients with image-space semantic supervision. Given a pre-trained $S(\cdot)$ and either $\{\X, \Y_{pseudo}\}$ (the self-supervised \methodname) or $\{\X, \Y\}$ (the supervised \methodname) as the target model training data, we invert $S(\cdot)$ to generated $\J_{T}$ as GT to directly supervise $C(\cdot)$. The $C(\cdot)$ output $\J$ is an image-like tensor, therefore $C(\cdot)$ can be trained with image-based losses against $J_{T}$ along with the source loss $\mathcal{L}_{S}$, leading to the Semantic Supervision (\textbf{SS}) loss,
\begin{equation}
\mathcal{L}_{SS}(\J, \J_{T})= SSIM(\J, \J_{T})+ L_1(\J, \J_{T}) + \mathcal{L}_{S},
\label{Eqn:SS}
\end{equation}  
where $SSIM(\cdot, \cdot)$ is the structural similarity loss and $L_{1}(\cdot,\cdot)$ is the L1-norm loss (Both losses provide \textbf{higher amplitudes gradients} than gradients back-propagated from $\mathcal{L}_{S}$). Due to the image-wise optimization nature of model inversion, each $\J_{T}$ over-fit different noise of the source model $S(\cdot)$. When training on all $\J_{T}$ samples, $C(\cdot)$ tends to produce averaged foreground semantics smoothing out sample-specific details that may be critical to detection.  

We propose a simple fix, called Decayed Semantic Supervision(\textbf{DSS}), to such a problem:
\begin{equation}
\mathcal{L}_{DSS}=\lambda_{DSS}(SSIM(\J, \J_{T})+ L_1(\J, \J_{T}))+ \mathcal{L}_{S},
\label{Eqn:DSS}
\end{equation}  
where $\lambda_{DSS}$ is a scalar that continuously decays with the increase of iterations (see supplementary materials).

\textbf{How DSS works?} The foreground semantics $\J_T$ only contain relatively clean foreground features reconstructed from $\Y$, but the background features are random due to the image-wise optimization of SMI. When training over all the data, using $\J_T$ throughout all training iterations will contaminate the target model with random background noise. By smoothly decaying lambda\_DSS, $\J_{T}$ provides strong gradients/supervision on foreground features in the early iterations, then the gradients from the source model losses $\mathcal{L}_{S}$ amend the acute overall details in later iterations. The effectiveness of $\textbf{DSS}$ is qualitatively shown in Figure~\ref{fig:MAC_Train}(b) and quantitatively evaluated in Table~\ref{table:Ablation}. 


\subsubsection{Skipped Inverted Attention (SIA)}
The Skipped Inverted Attention (SIA) is applied to amplify and balance the gradients from $\mathcal{L}_{S}$.
The strong gradients at the high-level layers of $S(\cdot)$ (see ``Avg. Gradients of Res50-p4to6
" in Figure~\ref{fig:MAC_Train}(a) of a ResNet-FPN-MaskRCNN $S(\cdot)$ module) does not propagate into strong ``Gradients at $\J$". 

To address this issue, we generate a 2D inverted attention mask $\A$ from the above gradients $\G$ and skip the earlier ResNet layers backward to supervise $C(\cdot)$. Formally, from $\G\in \Re^{width \times height\times channel}$, $\A(\G)\in \Re^{width \times height}$ is computed as,

\begin{equation}
    \mathbf{a}(i) = \begin{cases} 0, \quad \textnormal{if}\quad \sum_{j=1}^{channel} g(i,j) \geq q  \\
        1, \quad \textnormal{otherwise}
        \end{cases}
\end{equation}
where $g_p$ is a scalar of the $(100-p)$\textsuperscript{th} percentile of $\sum_{j=1}^{channel}\G(:,j)$, ($i\in [1, \cdots, width\times height]$). Low $\G$ and high $\A$ values denote under-represented regions by $S(\cdot)$ marked by white pixels in Figure~\ref{fig:MAC_Train}(b)``Inverted Attention $\A$ in SIA". Forwarding the element-wise masked feature $\A \odot \J$ through $S(\cdot)$, $T(\cdot)$ is updated by the \textbf{SIA} loss 
\begin{equation}
\mathcal{L}_{SIA}= \mathcal{L}_{S}(S(\A \odot \J), \Y).
\label{Eqn:Skipped}
\end{equation} 
Training with the $\A \odot \J$-induced loss $\mathcal{L}_{SIA}$, $C(\cdot)$ is forced to balance feature learning in all regions (See the strong foreground gradients at $\J$ in Figure~\ref{fig:MAC_Train}(c)). 


\hfill

In summary, the \methodname~Training procedure consists of initializing $T(\cdot)$ with \textbf{(1)} and updating $T(\cdot)$ with \textbf{(2-3)}. Figure~\ref{fig:MAC_Train}(c) shows that MAC training produces foreground-focused features in $\J$, stronger gradients at $\J$, and better instance mask detection than the \textit{Na\"ive} training in Figure~\ref{fig:MAC_Train}(a).

\section{Experiments} \label{exp}

We use the following keywords throughout this section. \textbf{``Standard"} refers to the source model training strategy (backbone pretrained on Imagenet and $S(\cdot)$ fully updated with $\mathcal{L}_{S}$). \noindent\textbf{``\methodname~training"} refers to our techniques. \noindent\textbf{``\methodname-Self-supervised":} Training on Pseudo-GT generated by inference $S(\cdot)$ with the source inputs of the target-source pairs in the target-modality training set. \noindent\textbf{``\methodname-Supervised":} Training on manually annotated GT of the target-modality training set. \noindent\textbf{``\methodname-Semi-Supervised":} Training on Pseudo-GT and a subset of manually annotated GT. 
All implementation details are described in the supplementary materials.  

\begin{table*}[!htb]
\centering
\fontsize{8}{9}\selectfont
\setlength{\tabcolsep}{1pt}
\begin{tabular}{l||c|c||c|c||c|c}
\hline
\multirow{2}{*}{Model (on \textbf{x101-GT})}        & \multirow{2}{*}{Input} & \multirow{2}{*}{Training Strategy} & Box  & Mask & Target-Modality & Inference\\
       &  &  & mAP $\uparrow$  & mAP $\uparrow$  &Annot. $(\%)\downarrow$ & Flops$|$\#Para.\\ \hline \hline
\textbf{\textit{Source models}} $S(\cdot)$& & & &&& \\
R50-FPN-MaskRCNN & RGB & Coco-pretrain~\cite{wu2019detectron2} + Standard & 82.36 & 87.94 &-& 61.42G$|$44.30M\\
 \hline
\textbf{\textit{Target models baselines}}& & & &&& \\
PiW~\cite{Wanghuang2019}$|S(\cdot)$ & CSI& Source Init. $S(\cdot)$+ Standard  & 59.86 & 45.08 &100&62.27G$|$45.0M \\
[3pt]
Wi2Vi~\cite{KefayatiPourahmadi2020}$|S(\cdot)$  & CSI & RGB-FG-pretrained Wi2Vi~\cite{KefayatiPourahmadi2020}& 0.12 & 0.09 &100 & 63.22G$|$49.82M \\
Wi2Vi~\cite{KefayatiPourahmadi2020}$|S(\cdot)$ & CSI & Source Init. $S(\cdot)$+Standard& 68.12 & 54.79 &100 & 63.22G$|$49.82M \\

\textbf{\textit{Ours target models $C(\cdot)|S(\cdot)$}}  &&& &&&\\
\methodname-Self-supervised &CSI &\methodname~training & 71.21 & 63.67 & \textbf{0}&70.10G$|$51.34M \\ 
\methodname-Semi-Supervised  &CSI &\methodname~training & 74.65 & 65.86 & 10 &70.10G$|$51.34M \\
\methodname-Supervised &CSI &\methodname~training & \textbf{77.38} & \textbf{66.49} & 100&70.10G$|$51.34M \\
\hline
\end{tabular}
\caption{WiFi CSI-input model results on the Person-in-Wifi(PiW) dataset (all $16$ layouts). All models were trained and evaluated against GTs generated by X101-FPN-MaskRCNN($\times 3$) in Detectron2 model zoo. The best target model metrics are in bold.}
\label{table:WiFi2Source}
\end{table*}

\begin{figure*}[!htb]
\centering
\includegraphics[width=.99\linewidth]{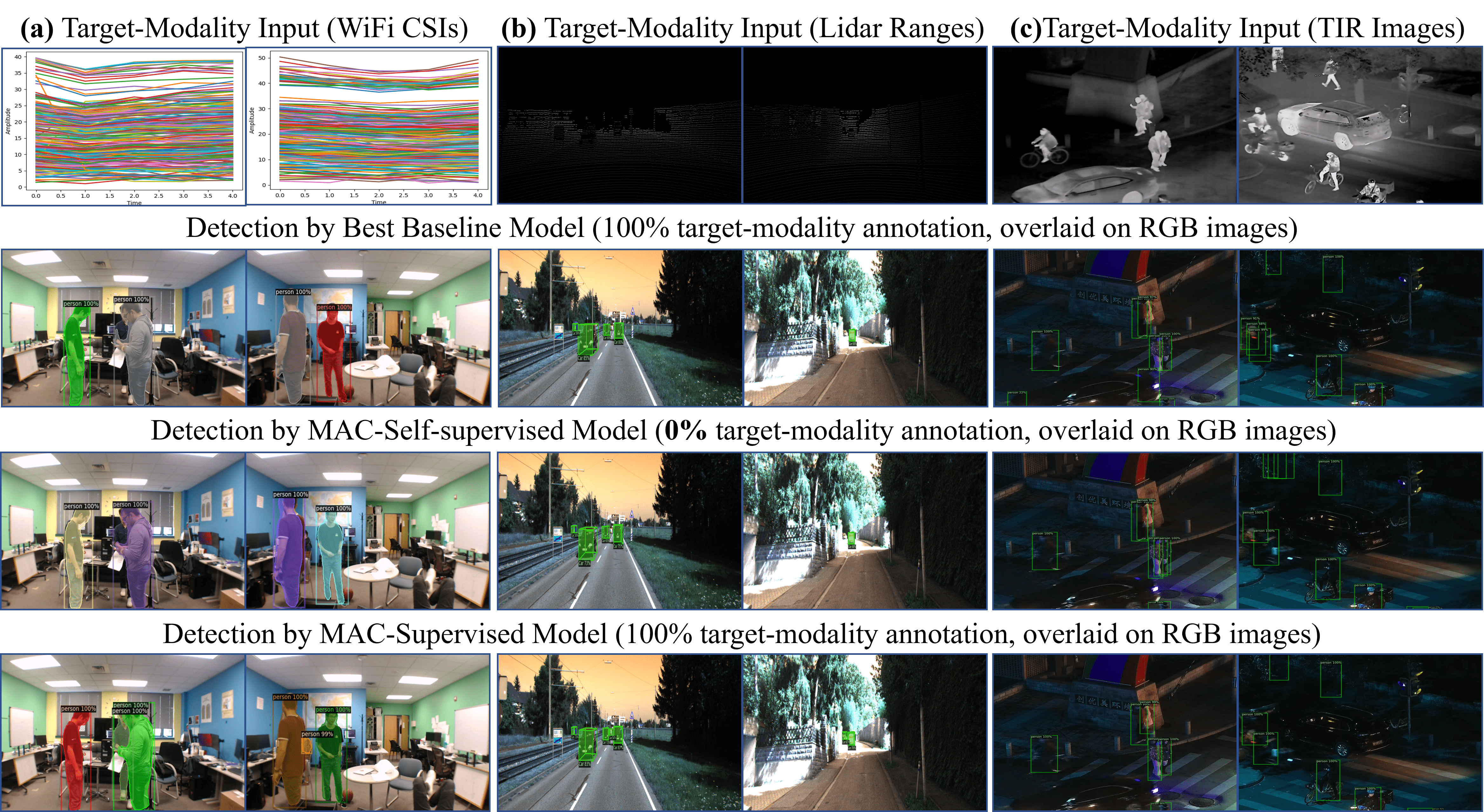}
\caption{Qualitative results on three target-modality inputs \textbf{(a)} WiFi CSIs (amplitude sequences corresponding to each frame), \textbf{(b)} Lidar Ranges (sparse depth points projected on the pixel grid) and \textbf{(c)} Thermal InFraRed (TIR) images.}
\label{fig:Q_results}
\end{figure*}

\subsection{WiFi-input Target Model} 
In Table~\ref{table:WiFi2Source}, we build WiFi-input models upon an RGB-input MaskRCNN model(source model) on the Person-in-Wifi(\textbf{PiW})~\cite{Wanghuang2019} dataset\footnote{https://www.donghuang-research.com/publications}. The PiW dataset contains synchronized RGB videos (20FPS) and Channel State Information(CSI) sequences (100Hz) of the Wifi signal. There are 16 indoor layouts 16  with multiple persons captured. One RGB frame (resized to $[3, 
640, 384]$) corresponds to a CSI tensor ([CSI\_samples, transmitters, receivers, sub-carriers]=$[5, 3,3, 30]$). The model in~\cite{Wanghuang2019} only produces image-wise semantic mask and body-joint heatmaps and cannot be directly compared on person detection metrics. No common ground truth was annotated to compare the RGB-input model and WiFi-input model.

To overcome these drawbacks, we propose the following setting: \textbf{(1)} All models are trained on and evaluated against a common ground truth \textbf{X101-GT}, which is generated by an MS-COCO-pre-trained ResNeXt101-FPN-MaskRCNN-32x8d(x3) model in Detectron2~\cite{wu2019detectron2} on RGB inputs. \textbf{(2)} Source model $S(\cdot)$: an RGB-input R50-FPN-MaskRCNN pre-trained on MS-COCO and fine-tuned on the PiW data. \textbf{(3)} Target model baseline (``PiW$|S(\cdot)$" and ``Wi2Vi$|S(\cdot)$"): composed by adding the CSI-to-RGB modules of PiW~\cite{Wanghuang2019} and Wi2Vi~\cite{KefayatiPourahmadi2020} to $S(\cdot)$ respectively. We train the target baselines by randomly initializing their CSI-to-RGB modules, initializing $S(\cdot)$ with source model weights, and training with the ``Standard" strategy. Pre-training the Wi2Vi module to synthesize the foreground-cropped images~\cite{KefayatiPourahmadi2020}, denoted by ``RGB-FG-pre-trained WiVi", does not work well on the multi-person and multi-layout PiW data.

``Our target models $C(\cdot)|S(\cdot)$" were trained by ``MAC" under three configurations: ``\methodname-Self-supervised" produces box and mask mAP of [$71.21, 63.67$] using \textbf{0\%} target-modality annotation, which outperforms the best target baseline metrics [$68.12, 54,79$] on 100\% target-modality annotation. This shows that \methodname~effectively transferred priors from the strong RGB-input model to the CSI-input model. Compared to the best baseline(Wi2Vi), the overheads of \methodname~models are 10\% in Flops and 3\% in \#Para. ``\methodname-Semi-supervised" and ``\methodname-Supervised" outperforms all other target models using $10$\% and 100\% target-modality annotation, respectively. Figure~\ref{fig:Q_results}(a) visualizes the results. 
 
\begin{table*}[!htb]
\centering
\fontsize{8}{9}\selectfont
\setlength{\tabcolsep}{1pt}
\begin{tabular}{l||c|c||c|c||c|c}
\hline
\multirow{2}{*}{Models}        & \multirow{2}{*}{Input} & \multirow{2}{*}{Training} & Car BEV AP$|_{40}$ & Car 3D-Bbox AP$|_{40}$  & Target-Modality & Inference\\
       &  &  & [Easy, Med., Hard]   $\uparrow$ & [Easy, Med., Hard]  $\uparrow$ &Annot. $(\%)\downarrow$ & Flops$|$\#Para. \\\hline \hline
\textbf{\textit{Source models}} $S(\cdot)$&  & &&&& \\
DD3D-DLA34~\cite{park2021dd3d}(github)& RGB & Standard & [31.7, 24.4, 21.7] & [22.6, 17.0, 14.9] & - & 109.9G$|$25.6M\\
\hline
\textbf{\textit{Target model baselines}}& &&  &&& \\
DD3D-DLA34  & Range & Standard & [40.7, 25.4, 22.0]  & [29.4, 17.9, 14.9] & $100$ & 109.9G$|$25.6M\\

\textbf{\textit{Ours target models $C(\cdot)|S(\cdot)$}}  &&& &&&\\
\methodname-Self-supervised &Range &\methodname & [41.5, 26.1, 23.2] & [30.2, 18.2, 15.4] & \textbf{0}&123.6G$|$28.1M \\
\methodname-Semi-Supervised  &Range &\methodname & [43.6, 27.3, 25.5]& [32.1, 19.8, 16.8] & 10&123.6G$|$28.1M \\
\methodname-Supervised &Range &\methodname & \textbf{[46.3, 33.4, 30.9]} & \textbf{[35.4, 21.8, 19.9]} & $100$&123.6G$|$28.1M \\

\hline
\end{tabular}
\caption{Lidar Range-input model results on the KITTI validation set. Metrics (Car metrics only) are evaluated on the frontal sector of Lidar scans matching the RGB Camera 2 field-of-view. The best target model metrics are in bold.}
\label{table:Lidar2Source}
\end{table*}

\subsection{Lidar-input Target Model}
We build Lidar Range-input models upon an RGB-input DD3D model~\cite{park2021dd3d} on the Kitti-3D dataset~\cite{Geiger2012CVPR}. Since there is no $360$-degree RGB coverage that matches the 360-degree Lidar scans, we only evaluate results on the \textbf{frontal-view sector} of the Lidar scans\footnote{LaserNet~\cite{Meyer2019LaserNet}, RangeDet~\cite{Fan_2021_ICCV} only reported results on 360-degree Range-inputs with no pre-trained model released to evaluate the frontal sector range data.} overlapped the RGB Camera\#2 field-of-view. The 32-beam Lidar scans at 1-degree horizontal resolution, creating $32\times90$ points in the frontal-view sector, which are then projected to the $384\times1270$ pixel grid to match the RGB 
pixels. Following~\cite{Fan_2021_ICCV,Meyer2019LaserNet}, the missing range pixels are filled by a fixed depth value of $80$ (meters). Following~\cite{park2021dd3d}, we report AP$|_{40}$ \cite{Simonelli2020} computed on the training$|$testing split of $3712|3769$ samples. As shown in Figure~\ref{fig:Q_results}(b), the range image has very sparse depth pixels with no visual appearance. 

In Table~\ref{table:Lidar2Source}, the ``Target model baseline", directly training a DD3D on Range-inputs, produces higher metrics than the RGB-input Source models, which indicates the advantage of the Lidar over the RGB camera. It appears that the weak RGB-input model cannot provide good prior knowledge to develop the Lidar-input model. However, the \methodname-Self-supervised target model, with \textbf{0\%} target-modality annotation, still outperforms the target baseline trained on 100\% annotation. Trained on 10\% and 100\% target-modality annotations respectively, our \methodname-Self-Supervised and \methodname-Supervised target models easily outperform all other models in all AP$|_{40}$ metrics. Figure~\ref{fig:Q_results}(b) visualizes the results.

\subsection{Infrared-input Target Model}
On the LLVIP dataset~\cite{jia2021llvip}~\footnote{https://github.com/bupt-ai-cz/LLVIP}, we show how \methodname~works when the source modality(RGB) contains far less information than the target modality(Thermal InfRared (TIR)) under low-light vision. We used the official training/testing split containing 15488 RGB-TIR pairs with manually labeled 2D bounding boxes.

In Table~\ref{table:TIR2Source}, we used the R50-FPN-FastRCNN (input size 1024$\times$1280) for both the source model (RGB input) and target model baseline (TIR input). Both models were pre-trained on MS-COCO and fine-tuned on the RGB and TIR inputs respectively. The \methodname~target models are trained with the \methodname~training algorithm. The source model produces Bbox Average Precision(Box AP) of $43.83$, which is clearly inferior to the target model baseline (Box AP $55.58$), therefore may not provide strong priors or correct Pseudo labels for the target model. However, the \methodname-Self-supervised target model still gets Box AP of $55.63$ using \textbf{0\%} target-modality annotation, which is comparable to the target baseline trained on 100\% annotation. With only 10\% annotation, the \methodname-Semi-supervised model easily outperforms the target baseline. With 100\% annotation, the \methodname-supervised target model outperforms all other models. Figure~\ref{fig:Q_results}(c) visualizes the results. In this experiment, the higher \methodname~overheads are due to the large input size, which could be reduced by concatenating $C(\cdot)$ with the smaller feature maps of $S(\cdot)$. 

\begin{table}[!htb]
\centering
\fontsize{8}{9}\selectfont
\setlength{\tabcolsep}{1pt}
\begin{tabular}{l||c|c||c||c|c}
\hline
\multirow{2}{*}{Models}        & \multirow{2}{*}{Input} & \multirow{2}{*}{Training} & Box  & Target-Modality & Inference\\
       &  &  & AP $\uparrow$   &Annot. $(\%)\downarrow$ & Flops$|$\#Para.\\\hline \hline
\textbf{\textit{Source model $S(\cdot)$}} &  & &&& \\

R50-FPN-FasterRCNN  & RGB & Standard &43.83 &-& 255G$|$41.70M\\
 \hline
\textbf{\textit{Target model baseline}}&  & &&& \\

R50-FPN-FasterRCNN  & TIR & Standard &55.58 &$100$ & 255G$|$41.70M\\

\textbf{\textit{Ours target models $C(\cdot)|S(\cdot)$}}  && &&&\\
\methodname-Self-supervised&TIR &\methodname & 55.63  & \textbf{0}& 302G$|$44.35M \\
\methodname-Semi-Supervised  &TIR &\methodname & 57.08 & $10$&302G$|$44.35M \\
\methodname-Supervised &TIR &\methodname & \textbf{58.05}  & $100$&302G$|$44.35M \\
\hline
\end{tabular}
\caption{Thermal InfRared TIR-input model results on the LLVIP dataset. Best in bold.}
\label{table:TIR2Source}
\end{table}

\subsection{Ablation Study} In Table~\ref{table:Ablation}, we conduct an ablation study of the \methodname~training techniques on the ``2018\_10\_17\_2" subset of the PiW dataset. We start with the baseline training strategy: ``RandInit+$\mathcal{L}_{S}$", and add \methodname~or alternative techniques in four groups. The cumulative relation among groups is denoted by their indents of ``$+$"s. Each group is added upon its previous \methodname~techniques. For instance, ``Feature-based KD..." and ``Source Init. $S(\cdot)$..." are both added upon ``FSR pretrained $C(\cdot)$...". 

\begin{table}[!htb]
\centering
\fontsize{8}{9}\selectfont
\setlength{\tabcolsep}{1pt}
\begin{tabular}{l||c|c||c}
\hline
\multirow{2}{*}{Training Strategies on $C(\cdot)|S(\cdot)$}        &  Box  & Mask & Target-Modality \\
           & AP $\uparrow$  & AP $\uparrow$  & Annot. (\%)$\downarrow$ \\\hline \hline


 Rand. Init.+Standard (Baseline)& 75.28 & 66.26 & $100$  \\

\hline
\hline
\textbf{\textit{MAC (bold rows) vs. Alternative techniques}}  && & \\ 
+ w/o FSR-pretrained $C(\cdot)$& 76.69 & 67,03 & $100$  \\
\textbf{+ FSR-pretrained $C(\cdot)$} & \textbf{78.83}& \textbf{68.73}& $100$  \\
\hline
 
\hspace{0.1cm}+ Feature-based KD~\cite{ZhangMa2021} from $S(\cdot)$& 70.90 & 54.43& $100$  \\ 
\hspace{0.1cm}\textbf{+ Source Init. $S(\cdot)$ and two-stage update}  & \textbf{80.36} & \textbf{70.55} & $100$  \\

\hline
\hspace{0.2cm}+ SS loss $\mathcal{L}_{SS}$ (Eq.(\ref{Eqn:SS})) & 81.03& 71.09& $100$  \\
\hspace{0.2cm}\textbf{+ DSS loss $\mathcal{L}_{DSS}$} (Eq.(\ref{Eqn:DSS}))  & \textbf{81.60}  & \textbf{71.62} & $100$  \\
\hline
\hspace{0.3cm}+ RSC~\cite{huangRSC2020} on the $C(\cdot)$ output layer  &  80.94  &71.21& $100$  \\
\hspace{0.3cm}+ RSC~\cite{huangRSC2020} on the Res-50-p5 layer &   81.89  & 71.92& $100$  \\
\hspace{0.3cm}\textbf{+ SIA}(Eq.(\ref{Eqn:Skipped}) & \textbf{82.15} & \textbf{72.33}& $100$ \\

\hline
MAC-Self-supervised & 75.14 & 65.96& \textbf{0}  \\
\hline

\end{tabular}
\caption{Ablation study of training techniques on \methodname~target model $C(\cdot)|S(\cdot)$ on the ``2018\_10\_17\_2" subset of PiW. The cumulative relation among groups of ``\methodname~vs.Alternative techniques" are denoted by their indents of ``$+$"s. Each group of techniques is added upon the \methodname~techniques (\textbf{bold rows}) of the previous group. }
\label{table:Ablation}
\end{table}

Within each compared group, the \methodname~technique (bold rows) produces better metrics than their alternative counterparts. ``FSR-pretrained $C(\cdot)$" and ``Source Init. $S(\cdot)$ and two-stage update" provide better priors to the target model than``w/o FSR" and ``Feature-based KD". SIA is better than RSC~\cite{huangRSC2020} which computes and applies attention on the same layer (``the C(·) output layer" or ``the Res-50-p5 layer"). Using 100\% target-modality annotation, our final model (after applying SIA) produces [82.15, 72.33], which is significantly better than the baseline ([75.28, 66.26]). Trained on the Pseudo GT generated by $S(\cdot)$, MAC-Self-supervised produces [75.14, 65.96] comparable to the baseline that is trained on $100\%$ manual target-modality annotation. Besides using the same number of pseudo images as the real images, we also tried fewer (0, 1/2) pseudo images and got [77.13,68.25] and [79.83,70.12] respectively. We also tried vanilla VAE as a calibrator and only got [0.5, 0.3] due to the VAE's inferior fine-grain reconstruction ability than VQVAE.

In summary, DSS is proven better than SS, and also improves upon ``Source Init. S(·) and two-stage update". SIA is proven better than two RSC variants, and also improves upon ``DSS loss LDSS (Eq.(3))". ``MAC-Self-supervised", that uses 0 manual annotations, has comparable performance as ``Baseline" that uses 100\% manual annotations. Using 100\% annotations, ``MAC-Supervised" is significantly better than ``Baseline".

\section{Discussion}
\textbf{Modality Calibration vs. Domain Adaption:} By definition\footnote{\url{https://en.wikipedia.org/wiki/Domain_adaptation}}, domain adaptation mainly addresses the shift of data sampling distribution. Our modality calibration problem mainly addresses the change in the spatial, temporal, and physical nature of the inputs. Although some work consider image translation as both modality calibration and domain adaption, our work applies to any non-image modalities such as WiFi signals. Instead of pursuing realistic pixels in image domain, we ONLY generate foreground-associated features contributing to the detection tasks, which aims to reduce the efforts in DNN design and data annotation. Moreover, for simplicity, this work is presented under the assumption that the source and target modality data are sampled from the same foreground/background distributions, such that MAC is only focused on reducing the discrepancy between modalities. 

\textbf{Pros. and Cons. of~\methodname:} By adding a calibrator to a pre-trained RGB perception model, MAC introduces an efficient pipeline for switching input modalities of DNNs, while bringing back the old deep learning challenge: the vanishing gradient problem ~\cite{Bengio1994,GlorotBengio2010}. To address this challenge, multiple MAC training techniques, FSR, DSS, and SIA, are developed to generate strong and dense gradients.

\textbf{MAC-Self-supervised vs. MAC-Supervised:} In ``MAC-Self-supervised" (see the cases of ``Target-Modality Annot." $0\%$ in Table~\ref{table:WiFi2Source}-\ref{table:Ablation}) the GTs for training the MAC target model are the pseudo annotations produced by inferencing a pre-trained source model. There is NO manual GT for the target modality. ``MAC-Supervised" uses the real manually annotated GTs (see the case of ``MAC-Supervised" with ``Target-Modality Annot." $100\%$ in Table~\ref{table:WiFi2Source}-\ref{table:Ablation}). This presents us with options to balance the performance and the effort of annual data annotation.

\textbf{General use of MAC: } The same FSR, DDS, and SIA training techniques can be used to develop all non-RGB modality models. Only different calibrators need to be developed for different non-RGB modalities.

\section{Conclusions} \label{chapConclusions}

We proposed \methodname, an efficient pipeline for switching input modalities of DNNs. In training a target-modality model, \methodname~leverages the prior knowledge from the source-modality model and requires as few as \textbf{zero} target-modality annotations. The \methodname~components (FSR, DSS, SIA) could potentially be used to compose any cross-modality models, for instance, using DALL-E-mini as a calibrator to compose a text-input model.  

\textit{Potential negative social impact:} If the source model were trained on private RGB images, data-privacy concerns may arise from the source model inversion operation in \methodname. 

\section*{Appendix}

\subsection*{Implementation Details}

\textbf{WiFi-input Model} 
For the WiFi-input model, we use R50-FPN-MaskRCNN \cite{DBLP:journals/corr/HeGDG17} as our source model, and Wi2Vi \cite{KefayatiPourahmadi2020} + VQ-VAE2 \cite{VQVAE2019} as calibrator. We used two-level latent maps which are approximately 16x, 2x times smaller than the original image. The codebook size is set as 1024. The decay factor for Decayed Semantic Supervision $\lambda_{DSS}$ is set as 0.9999. The percentile threshold $p$ used in Skipped Inverted Attention is 0.1. 

We use 8 GPUs with batch size 64 for training. Following the R50-FPN-MaskRCNN training codes, the MAC target-input model is trained using the Adam optimizer for 150k iterations with a weight decay of 0.0001 and an initial learning rate of 0.0001. The training process is warmed up by a linear warm-up-scheduler with 0.001 factor for 1k iterations and then the learning rate is decayed by 10x times at 90k iterations.


\textbf{Lidar-input Model} 
For the Lidar-input model, we use DD3D \cite{park2021dd3d} as our source model and VQVAE2 as calibrator. We used three-level latent maps which are approximately 32x, 4x, 2x, times smaller than the original image. The codebook size is set as 4096. We use 8 GPUs with batch size 64 for training. 
The decay factor for Decayed Semantic Supervision $\lambda_{DSS}$ is set as 0.9995. The percentile threshold $p$ used in Skipped Inverted Attention is 0.1. 

Following DD3D training codes, the MAC target-input model is trained using SGD optimizer for 24k iterations with weight decay 0.0001, momentum 0.9, and an initial learning rate of 0.002. The training process is warmed up by a linear warm-up-scheduler with 0.001 factor for 2k iterations and then the learning rate is decayed by 10x times at 21.5k and 25k iterations. 

\textbf{Infrared-input Model} 
For the Infrared-input model, we use R50-FPN-FastRCNN as our source model and VQVAE2 as the calibrator. We used three-level latent maps which are approximately 32x, 4x, 2x, times smaller than the original image respectively. The codebook size is set as 4096. We use 8 GPUs with batch size 64 for training. The decay factor for Decayed Semantic Supervision $\lambda_{DSS}$ is set as 0.9999. The percentile threshold $p$ used in Skipped Inverted Attention is 0.1. 

Following the R50-FPN-FastRCNN training codes, the MAC target-input model is trained using Adam optimizer for 50k iterations with a weight decay of 0.0001 and an initial learning rate of 0.0001. The training process is warmed up by a linear warm-up-scheduler with 0.001 factor for 1k iterations and then the learning rate is decayed by 10x times at 30k iterations. 

\textbf{Training/inference Costs} Take the setting in the main paper Table 4 for example. Compared to training the [Calibrator|Source] model from scratch (Baseline in Table 4), MAC training introduces three main overheads: pre-computing Foreground Semantics $J$, pretraining $C(\cdot)$ in FSR, and SIA, which respectively cost around 200\%, 20\%, and 10\% of the baseline training time. Note that pre-computing Foreground Semantics $J$ is a one-time data pre-processing stage and does not join the training iteration of the target-modality model. The inference overhead is introduced only by the calibrator over the source model. In the main paper Table 1-3, the calibrators add 14\% 13\%, and 15\% of the source model inference time, respectively. 

\subsection*{Source Model Inversion Examples}
\begin{figure*}[!htb]
    \centering
    \includegraphics[width=\linewidth]{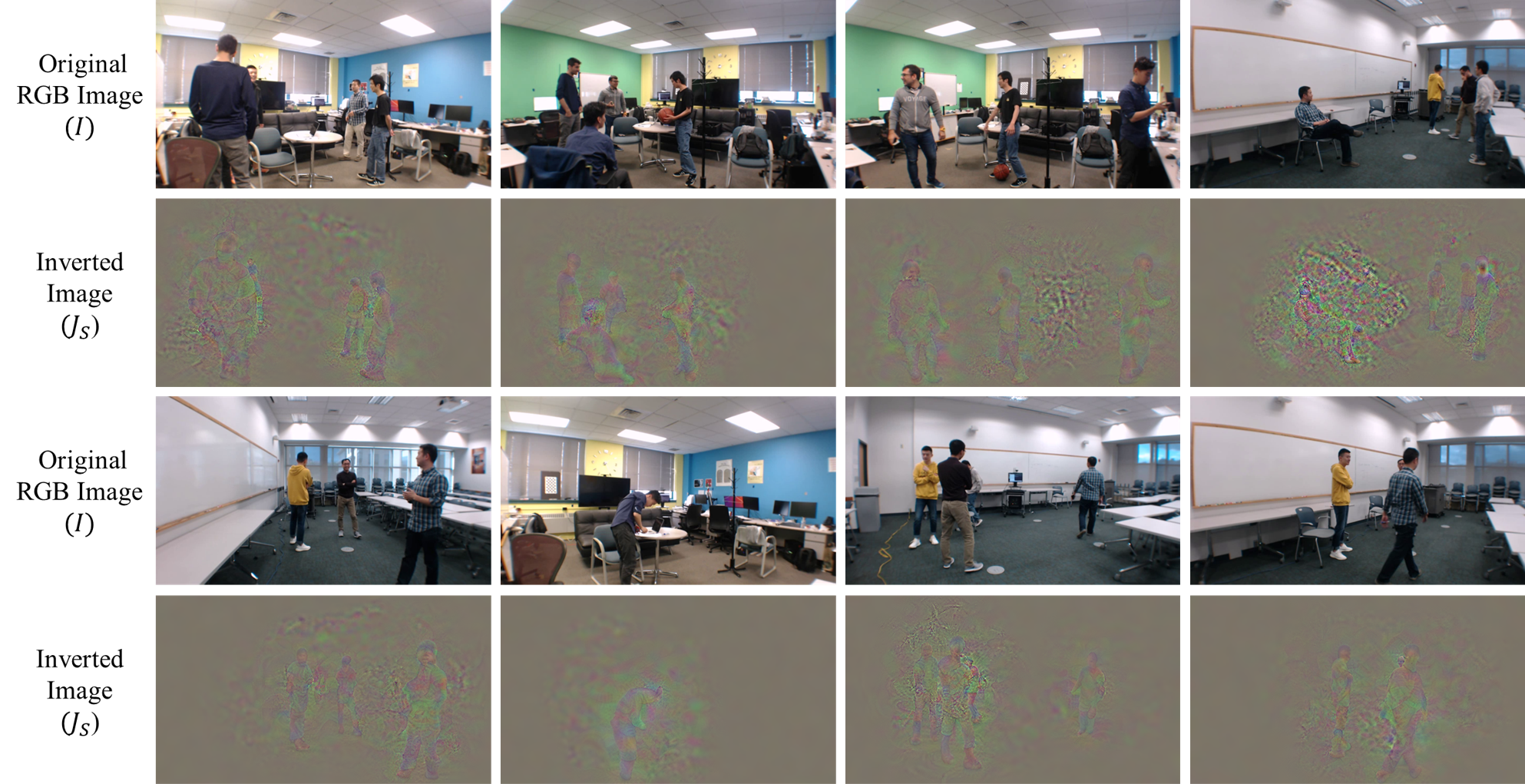}
    \caption{Model inversion examples of the R50-FPN-MaskRCNN source model in the ``WiFi-input Target Model" experiments}
    \label{fig:inv_wifi}
\end{figure*}

\begin{figure*}[!htb]
    \centering
    \includegraphics[width=\linewidth]{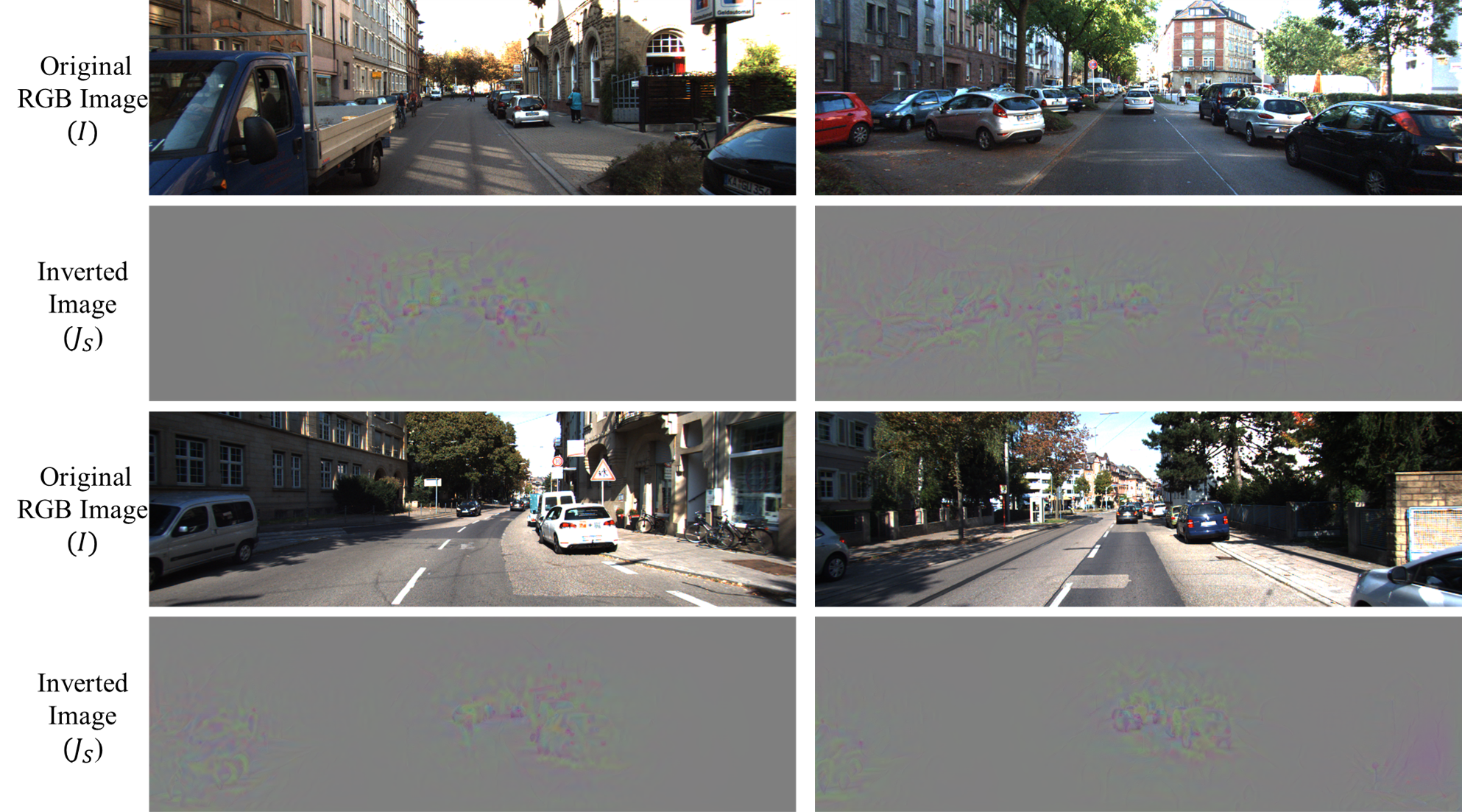}
    \caption{Model inversion examples of the DD3D source model in the ``Lidar-input Target Model" experiments.}
    \label{fig:inv_dd3d}
\end{figure*}

\begin{figure*}[!htb]
    \centering
    \includegraphics[width=\linewidth]{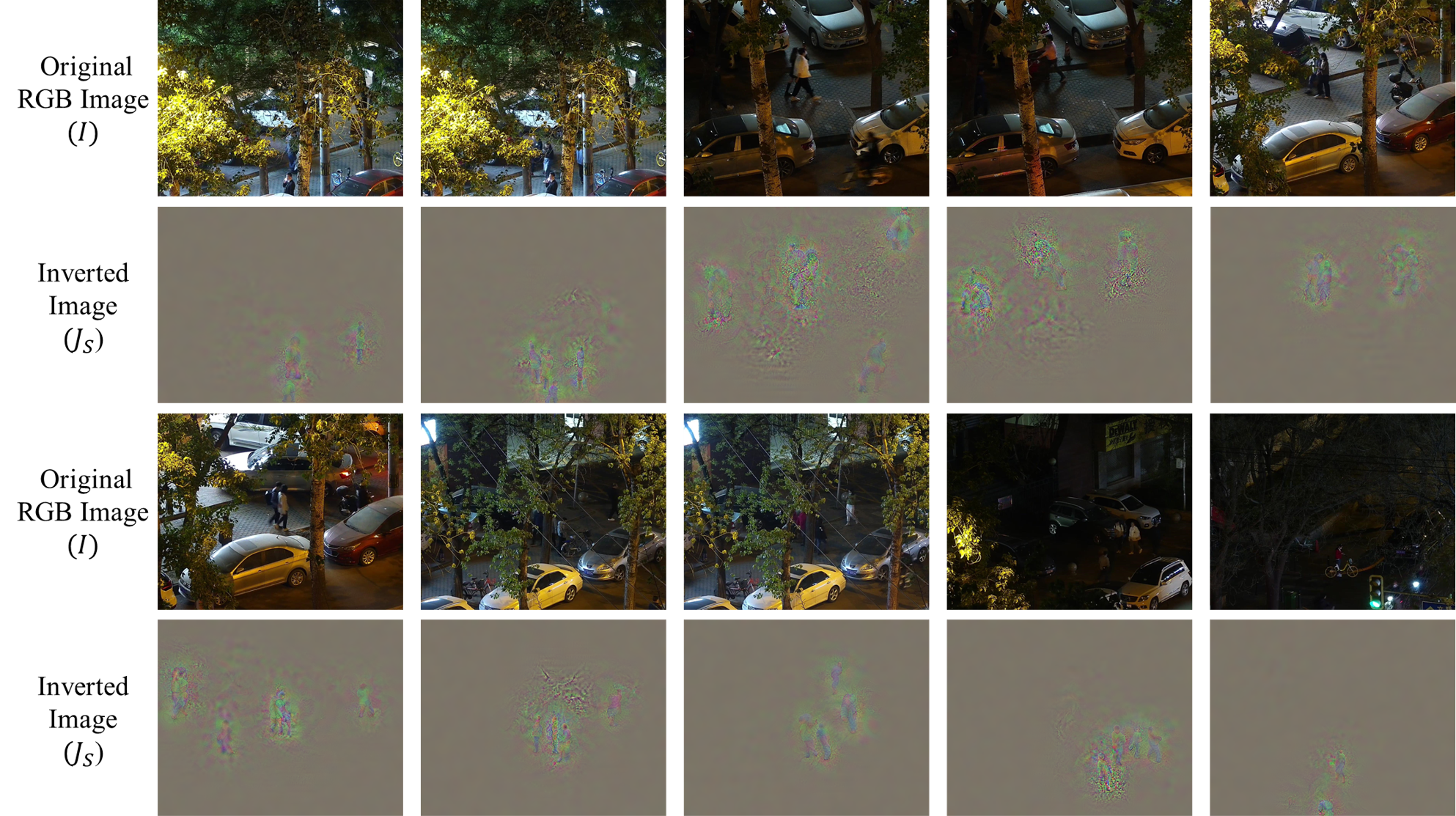}
    \caption{Model inversion examples of the R50-FPN-FasterRCNN source model in the ``Infrared-input Target Model" experiments.}
    \label{fig:inv_LLVIP}
\end{figure*}

Our Source Model Inversion(SMI) generate foreground semantics images $\J_{S}$. We show examples of SMI respectively on the three different source models in Figure \ref{fig:inv_wifi}, Figure \ref{fig:inv_dd3d} and Figure \ref{fig:inv_LLVIP}. In all cases, $\J_{S}$ shows clear foreground-sensitive features. 

\subsection*{Target-modality Qualitative Results}
\begin{figure*}[!htb]
    \centering
    \includegraphics[width=\linewidth]{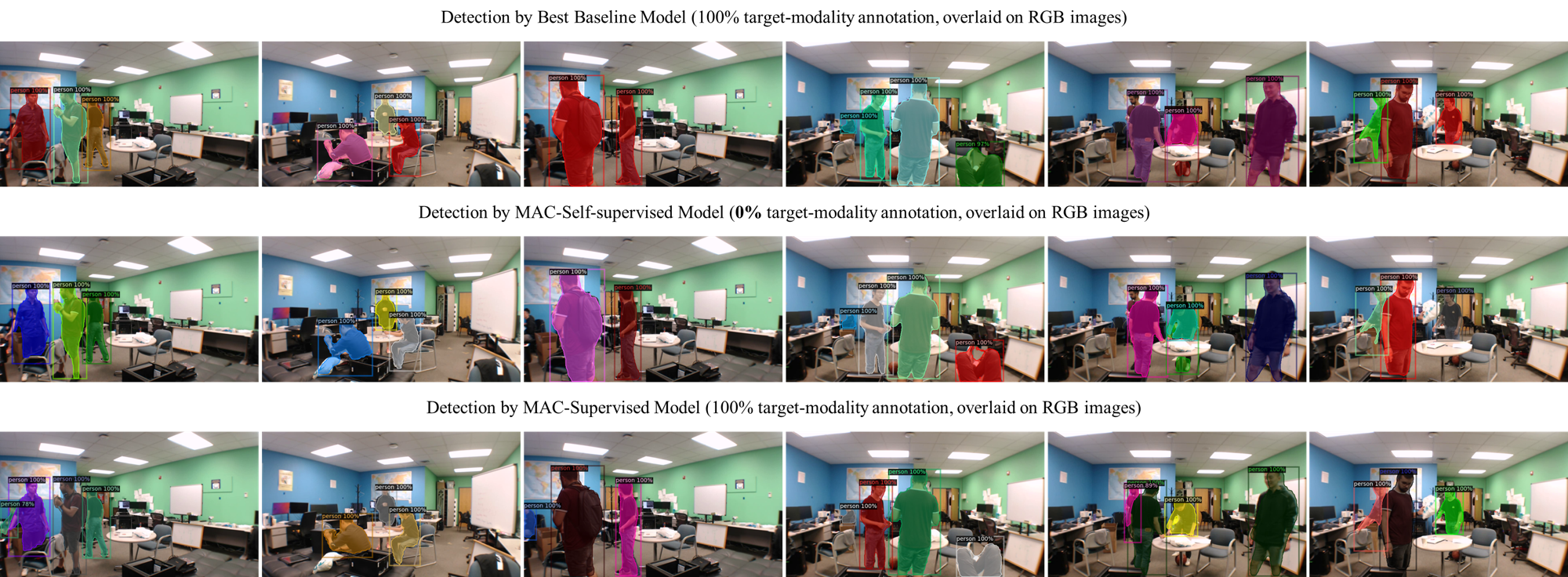}
    \caption{More Qualitative results on WiFi-input models (overlaid on RGB images).}
    \label{fig:wifi}
\end{figure*}

\begin{figure*}[!htb]
    \centering
    \includegraphics[width=\linewidth]{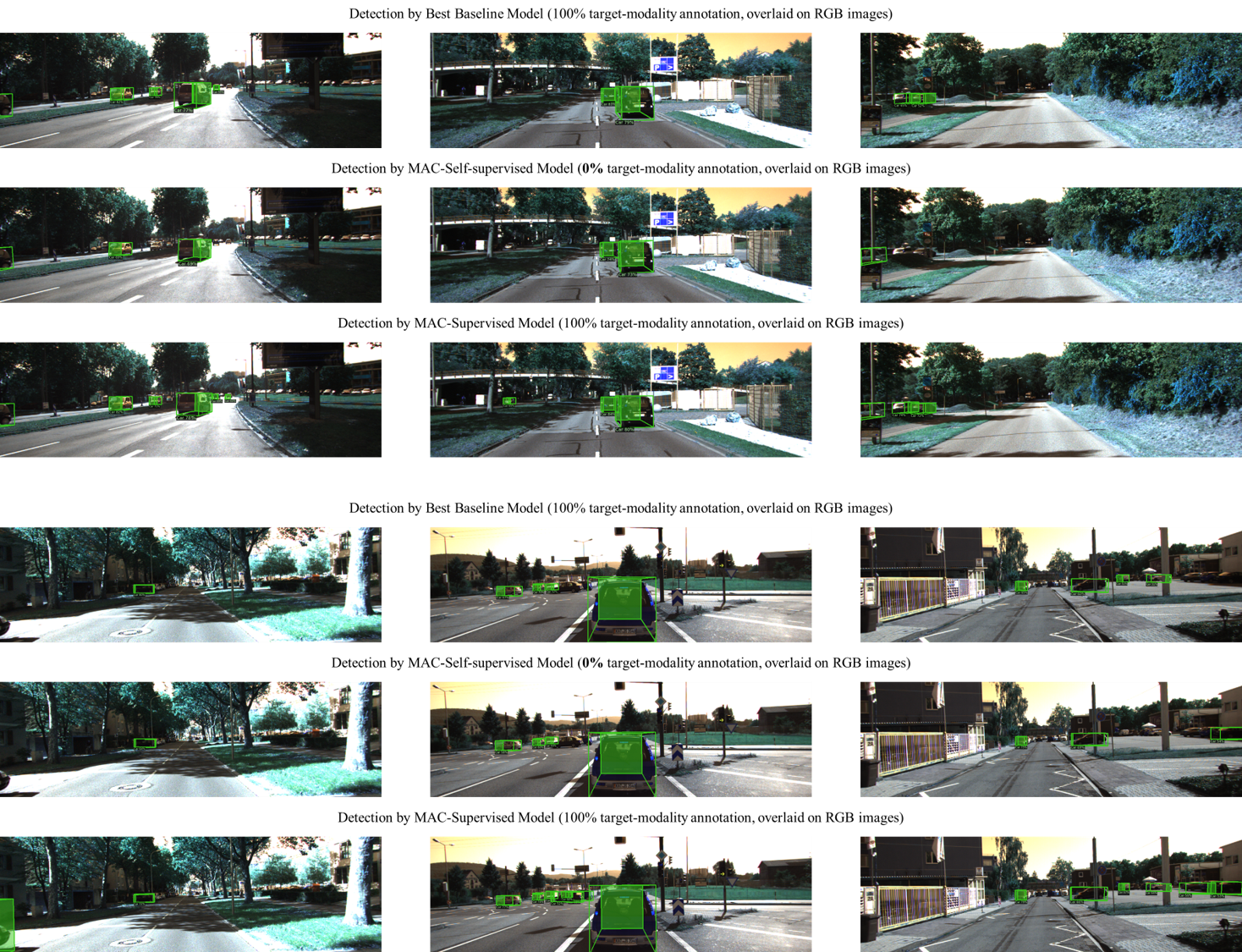}
    \caption{More Qualitative results on Lidar-input models (overlaid on RGB images).}
    \label{fig:dd3d}
\end{figure*}

\begin{figure*}[!htb]
    \centering
    \includegraphics[width=\linewidth]{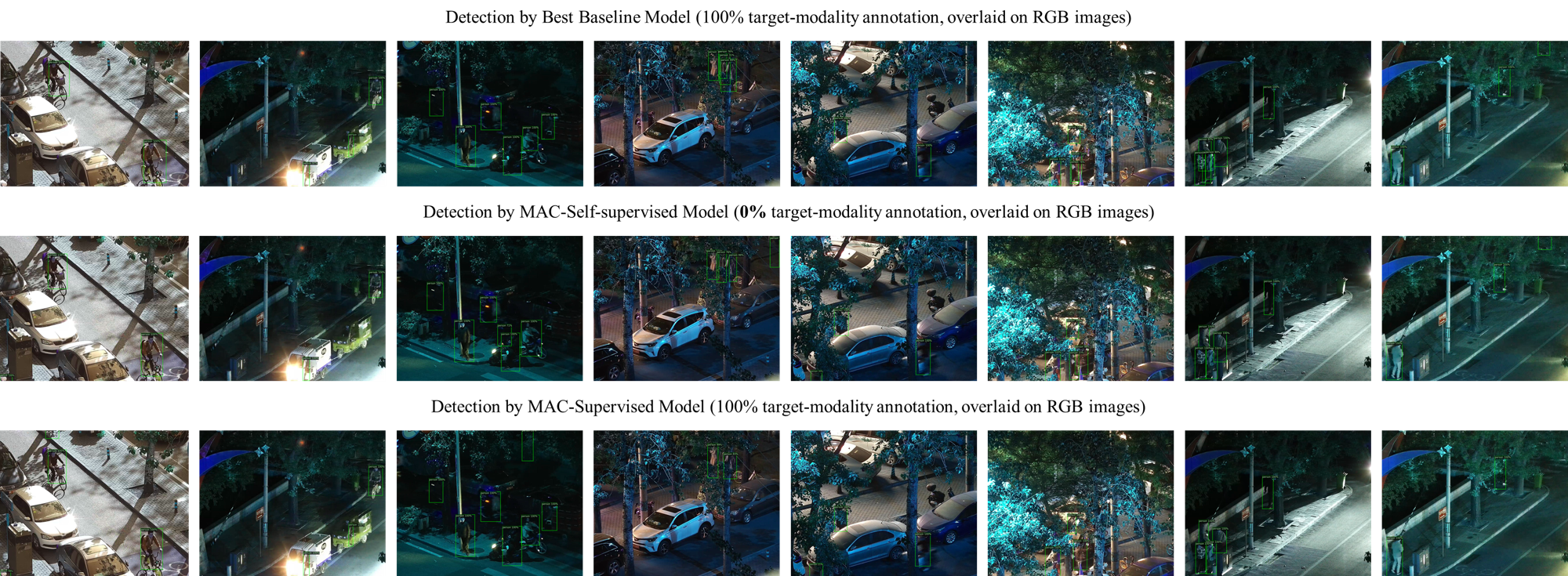}
    \caption{More Qualitative results on Infrared-input models (overlaid on RGB images).}
    \label{fig:LLVIP}
\end{figure*}

In Figure \ref{fig:wifi}, Figure \ref{fig:dd3d}, and Figure \ref{fig:LLVIP}, we report more qualitative results on three target-modality input models. In all cases, the MAC-Self-supervised models that are trained on 0\% target-modality annotations produce comparable or better than the best baseline models trained on 100\% annotations. Trained on the same 100\% target-modality annotations, the MAC-Supervised models clearly outperform the best baseline models. 

\bibliographystyle{unsrtnat}
\bibliography{references}  

\begin{thebibliography}{74}
\providecommand{\natexlab}[1]{#1}
\providecommand{\url}[1]{\texttt{#1}}
\expandafter\ifx\csname urlstyle\endcsname\relax
  \providecommand{\doi}[1]{doi: #1}\else
  \providecommand{\doi}{doi: \begingroup \urlstyle{rm}\Url}\fi

\bibitem[Zhao et~al.(2018)Zhao, Li, Abu~Alsheikh, Tian, Zhao, Torralba, and Katabi]{zhao2018through}
Mingmin Zhao, Tianhong Li, Mohammad Abu~Alsheikh, Yonglong Tian, Hang Zhao, Antonio Torralba, and Dina Katabi.
\newblock Through-wall human pose estimation using radio signals.
\newblock In \emph{CVPR}, pages 7356--7365, 2018.

\bibitem[Wang et~al.(2019{\natexlab{a}})Wang, Zhou, Panev, Han, and Huang]{Wanghuang2019}
Fei Wang, Sanping Zhou, Stanislav Panev, Jinsong Han, and Dong Huang.
\newblock Person-in-wifi: Fine-grained person perception using wifi.
\newblock In \emph{ICCV}, 2019{\natexlab{a}}.

\bibitem[Geiger et~al.(2012)Geiger, Lenz, and Urtasun]{Geiger2012CVPR}
Andreas Geiger, Philip Lenz, and Raquel Urtasun.
\newblock Are we ready for autonomous driving? the kitti vision benchmark suite.
\newblock In \emph{CVPR}, 2012.

\bibitem[Sun et~al.(2020)Sun, Kretzschmar, Dotiwalla, and et~al.]{Sun2020Waymo}
Pei Sun, Henrik Kretzschmar, Xerxes Dotiwalla, and et~al.
\newblock Scalability in perception for autonomous driving: Waymo open dataset.
\newblock In \emph{CVPR}, 2020.

\bibitem[Hwang et~al.(2015)Hwang, Park, Kim, Choi, and Kweon]{hwang2015multispectral}
Soonmin Hwang, Jaesik Park, Namil Kim, Yukyung Choi, and In~So Kweon.
\newblock Multispectral pedestrian detection: Benchmark dataset and baselines.
\newblock In \emph{CVPR}, 2015.

\bibitem[Jia et~al.(2021)Jia, Zhu, Li, Tang, and Zhou]{jia2021llvip}
Xinyu Jia, Chuang Zhu, Minzhen Li, Wenqi Tang, and Wenli Zhou.
\newblock Llvip: A visible-infrared paired dataset for low-light vision.
\newblock In \emph{Proceedings of the ICCV}, pages 3496--3504, 2021.

\bibitem[He et~al.(2017)He, Gkioxari, Doll{\'{a}}r, and Girshick]{DBLP:journals/corr/HeGDG17}
Kaiming He, Georgia Gkioxari, Piotr Doll{\'{a}}r, and Ross~B. Girshick.
\newblock Mask {R-CNN}.
\newblock \emph{CoRR}, abs/1703.06870, 2017.
\newblock URL \url{http://arxiv.org/abs/1703.06870}.

\bibitem[Jocher et~al.(2020)Jocher, Stoken, Borovec, NanoCode012, ChristopherSTAN, Changyu, Laughing, tkianai, Hogan, lorenzomammana, yxNONG, AlexWang1900, Diaconu, Marc, wanghaoyang0106, ml5ah, Doug, Ingham, Frederik, Guilhen, Hatovix, Poznanski, Fang, Yu, changyu98, Wang, Gupta, Akhtar, PetrDvoracek, and Rai]{glenn_jocher_2020_4154370}
Glenn Jocher, Alex Stoken, Jirka Borovec, NanoCode012, ChristopherSTAN, Liu Changyu, Laughing, tkianai, Adam Hogan, lorenzomammana, yxNONG, AlexWang1900, Laurentiu Diaconu, Marc, wanghaoyang0106, ml5ah, Doug, Francisco Ingham, Frederik, Guilhen, Hatovix, Jake Poznanski, Jiacong Fang, Lijun Yu, changyu98, Mingyu Wang, Naman Gupta, Osama Akhtar, PetrDvoracek, and Prashant Rai.
\newblock {ultralytics/yolov5: v3.1 - Bug Fixes and Performance Improvements}, October 2020.
\newblock URL \url{https://doi.org/10.5281/zenodo.4154370}.

\bibitem[Liu et~al.(2021)Liu, Lin, Cao, Hu, Wei, Zhang, Lin, and Guo]{liu2021Swin}
Ze~Liu, Yutong Lin, Yue Cao, Han Hu, Yixuan Wei, Zheng Zhang, Stephen Lin, and Baining Guo.
\newblock Swin transformer: Hierarchical vision transformer using shifted windows.
\newblock \emph{arXiv:2103.14030}, 2021.

\bibitem[Deng et~al.(2009)Deng, Dong, Socher, Li, Li, and Fei-Fei]{imagenet_cvpr09}
J.~Deng, W.~Dong, R.~Socher, L.-J. Li, K.~Li, and L.~Fei-Fei.
\newblock {ImageNet: A Large-Scale Hierarchical Image Database}.
\newblock In \emph{CVPR09}, 2009.

\bibitem[Lin et~al.(2014)Lin, Maire, Belongie, Hays, Perona, Ramanan, Dollár, and Zitnick]{ms_coco}
Tsung-Yi Lin, Michael Maire, Serge Belongie, James Hays, Pietro Perona, Deva Ramanan, Piotr Dollár, and C.~Lawrence Zitnick.
\newblock Microsoft coco: Common objects in context.
\newblock In \emph{ECCV}, Zürich, 2014.
\newblock URL \url{/se3/wp-content/uploads/2014/09/coco_eccv.pdf, http://mscoco.org}.

\bibitem[Kuznetsova et~al.(2020)Kuznetsova, Rom, Alldrin, Uijlings, Krasin, Pont-Tuset, Kamali, Popov, Malloci, Kolesnikov, Duerig, and Ferrari]{OpenImages}
Alina Kuznetsova, Hassan Rom, Neil Alldrin, Jasper Uijlings, Ivan Krasin, Jordi Pont-Tuset, Shahab Kamali, Stefan Popov, Matteo Malloci, Alexander Kolesnikov, Tom Duerig, and Vittorio Ferrari.
\newblock The open images dataset v4: Unified image classification, object detection, and visual relationship detection at scale.
\newblock \emph{IJCV}, 2020.

\bibitem[He et~al.(2018)He, Zhang, Zhang, Zhang, Xie, and Li]{he2018bag}
Tong He, Zhi Zhang, Hang Zhang, Zhongyue Zhang, Junyuan Xie, and Mu~Li.
\newblock Bag of tricks for image classification with convolutional neural networks.
\newblock \emph{arXiv:1812.01187}, 2018.

\bibitem[Zhang et~al.(2019)Zhang, He, Zhang, Zhang, Xie, and Li]{zhang2019bag}
Zhi Zhang, Tong He, Hang Zhang, Zhongyue Zhang, Junyuan Xie, and Mu~Li.
\newblock Bag of freebies for training object detection neural networks.
\newblock \emph{arXiv:1902.04103}, 2019.

\bibitem[Ghiasi et~al.(2020)Ghiasi, Cui, Srinivas, Qian, Lin, Cubuk, Le, and Zoph]{ghiasi2020simple}
Golnaz Ghiasi, Yin Cui, Aravind Srinivas, Rui Qian, Tsung-Yi Lin, Ekin~D Cubuk, Quoc~V Le, and Barret Zoph.
\newblock Simple copy-paste is a strong data augmentation method for instance segmentation.
\newblock \emph{arXiv:2012.07177}, 2020.

\bibitem[Wu et~al.(2019)Wu, Kirillov, Massa, Lo, and Girshick]{wu2019detectron2}
Yuxin Wu, Alexander Kirillov, Francisco Massa, Wan-Yen Lo, and Ross Girshick.
\newblock \url{Detectron2}, 2019.

\bibitem[Chen et~al.(2019)Chen, Wang, Pang, Cao, Xiong, Li, Sun, Feng, Liu, Xu, Zhang, Cheng, Zhu, Cheng, Zhao, Li, Lu, Zhu, Wu, Dai, Wang, Shi, Ouyang, Loy, and Lin]{mmdetection}
Kai Chen, Jiaqi Wang, Jiangmiao Pang, Yuhang Cao, Yu~Xiong, Xiaoxiao Li, Shuyang Sun, Wansen Feng, Ziwei Liu, Jiarui Xu, Zheng Zhang, Dazhi Cheng, Chenchen Zhu, Tianheng Cheng, Qijie Zhao, Buyu Li, Xin Lu, Rui Zhu, Yue Wu, Jifeng Dai, Jingdong Wang, Jianping Shi, Wanli Ouyang, Chen~Change Loy, and Dahua Lin.
\newblock {MMDetection}: Open mmlab detection toolbox and benchmark.
\newblock \emph{arXiv:1906.07155}, 2019.

\bibitem[Qi et~al.(2017)Qi, Yi, Su, and Guibas]{qi2017pointnetplusplus}
Charles~R Qi, Li~Yi, Hao Su, and Leonidas~J Guibas.
\newblock Pointnet++: Deep hierarchical feature learning on point sets in a metric space.
\newblock \emph{arXiv:1706.02413}, 2017.

\bibitem[Kato et~al.(2021)Kato, Fukushima, Murakami, Abeysekera, Iwasaki, Fujihashi, Watanabe, and Saruwatari]{KatoFukushima2021}
Sorachi Kato, Takeru Fukushima, T.~Murakami, H.~Abeysekera, Yusuke Iwasaki, T.~Fujihashi, Takashi Watanabe, and S.~Saruwatari.
\newblock Csi2image: Image reconstruction from channel state information using generative adversarial networks.
\newblock \emph{IEEE Access}, 9:\penalty0 47154--47168, 2021.

\bibitem[Kefayati et~al.(2020)Kefayati, Pourahmadi, and Aghaeinia]{KefayatiPourahmadi2020}
Mohammad~Hadi Kefayati, Vahid Pourahmadi, and Hassan Aghaeinia.
\newblock Wi2vi: Generating video frames from wifi csi samples.
\newblock \emph{IEEE Sensors Journal}, 20\penalty0 (19):\penalty0 11463--11473, 2020.

\bibitem[Drob(2021)]{Drob2021}
Michael Drob.
\newblock Rf pix2pix unsupervised wi-fi to video translation.
\newblock In \emph{arXiv:2102.09345}, 2021.

\bibitem[Milz et~al.(2019)Milz, Simon, Fischer, and Pöpperl]{MilzSimon2019}
Stefan Milz, Martin Simon, Kai Fischer, and Maximillian Pöpperl.
\newblock Points2pix: 3d point-cloud to image translation using conditional generative adversarial networks.
\newblock \emph{arXiv:1901.09280}, 2019.

\bibitem[Limmer and Lensch(2019)]{Limmer2019}
Matthias Limmer and Hendrik~P.A. Lensch.
\newblock Infrared colorization using deep convolutional neural networks.
\newblock \emph{arXiv:1604.02245}, 2019.

\bibitem[Rajendran et~al.(2019)Rajendran, Trongtirakul, Trongtirakul, Panetta, and Agaian]{Rajendran2019}
Rahul Rajendran, Thaweesak Trongtirakul, Thaweesak Trongtirakul, Karen Panetta, and Sos Agaian.
\newblock A pixel-based color transfer system to recolor nighttime imagery.
\newblock In \emph{Mobile Multimedia/Image Processing, Security, and Applications}, 2019.

\bibitem[Karras et~al.(2020)Karras, Laine, Aittala, Hellsten, Lehtinen, and Aila]{Karras2019stylegan2}
Tero Karras, Samuli Laine, Miika Aittala, Janne Hellsten, Jaakko Lehtinen, and Timo Aila.
\newblock Analyzing and improving the image quality of {StyleGAN}.
\newblock In \emph{Proc. CVPR}, 2020.

\bibitem[Depatla and Mostofi(2018)]{Depatla2018}
S.~Depatla and Y.~Mostofi.
\newblock Crowd counting through walls using wifi.
\newblock In \emph{Proceedings of IEEE International Conference on Pervasive Computing and Communications}, 2018.

\bibitem[Liu et~al.(2019{\natexlab{a}})Liu, Zhao, Xue, Chen, and Chen]{liu2019deepcount}
Shangqing Liu, Yanchao Zhao, Fanggang Xue, Bing Chen, and Xiang Chen.
\newblock Deepcount: Crowd counting with wifi via deep learning.
\newblock \emph{arXiv:1903.05316}, 2019{\natexlab{a}}.

\bibitem[Wang et~al.(2017{\natexlab{a}})Wang, Liu, Shahzad, Ling, and Lu]{WangLiu2017}
Wei Wang, Alex~X. Liu, Muhammad Shahzad, Kang Ling, and Sanglu Lu.
\newblock Device-free human activity recognition using commercial wifi devices.
\newblock \emph{IEEE Journal on Selected Areas in Communications}, 35\penalty0 (5):\penalty0 1118--1131, 2017{\natexlab{a}}.

\bibitem[Li et~al.(2019)Li, He, Chen, Fang, and Fang]{LiHe2019}
Heju Li, Xin He, Xukai Chen, Yinyin Fang, and Qun Fang.
\newblock Wi-motion: A robust human activity recognition using wifi signals.
\newblock \emph{IEEE Access}, 7:\penalty0 153287 -- 153299, 2019.

\bibitem[Qi et~al.(2016)Qi, Su, Mo, and Guibas]{qi2016pointnet}
Charles~R Qi, Hao Su, Kaichun Mo, and Leonidas~J Guibas.
\newblock Pointnet: Deep learning on point sets for 3d classification and segmentation.
\newblock \emph{arXiv:1612.00593}, 2016.

\bibitem[Shi et~al.(2019)Shi, Wang, and Li]{Shi_2019_CVPR}
Shaoshuai Shi, Xiaogang Wang, and Hongsheng Li.
\newblock Pointrcnn: 3d object proposal generation and detection from point cloud.
\newblock In \emph{CVPR}, June 2019.

\bibitem[Yan et~al.(2018)Yan, Mao, and Li]{yan2018second}
Yan Yan, Yuxing Mao, and Bo~Li.
\newblock Second: Sparsely embedded convolutional detection.
\newblock \emph{Sensors}, 2018.

\bibitem[Lang et~al.(2019)Lang, Vora, Caesar, Zhou, Yang, and Beijbom]{lang2019pointpillars}
Alex~H Lang, Sourabh Vora, Holger Caesar, Lubing Zhou, Jiong Yang, and Oscar Beijbom.
\newblock Pointpillars: Fast encoders for object detection from point clouds.
\newblock In \emph{CVPR}, pages 12697--12705, 2019.

\bibitem[Meyer et~al.(2019)Meyer, Laddha, Kee, VallespiGonzalez, and Wellington]{Meyer2019LaserNet}
Gregory~P Meyer, Ankit Laddha, Eric Kee, Carlos VallespiGonzalez, and Carl~K Wellington.
\newblock Lasernet: An efficient probabilistic 3d object detector for autonomous driving.
\newblock In \emph{CVPR}, page 12677–12686, 2019.

\bibitem[Fan et~al.(2021)Fan, Xiong, Wang, Wang, and Zhang]{Fan_2021_ICCV}
Lue Fan, Xuan Xiong, Feng Wang, Naiyan Wang, and ZhaoXiang Zhang.
\newblock Rangedet: In defense of range view for lidar-based 3d object detection.
\newblock In \emph{ICCV}, October 2021.

\bibitem[Murez et~al.(2018)Murez, Kolouri, Kriegman, Ramamoorthi, and Kim]{Murez2018}
Zak Murez, Soheil Kolouri, David Kriegman, Ravi Ramamoorthi, and Kyungnam Kim.
\newblock Image to image translation for domain adaptation.
\newblock In \emph{CVPR}, 2018.

\bibitem[Dou et~al.(2018)Dou, Ouyang, Chen, Chen, and Heng]{dou2018unsupervised}
Qi~Dou, Cheng Ouyang, Cheng Chen, Hao Chen, and Pheng-Ann Heng.
\newblock Unsupervised cross-modality domain adaptation of convnets for biomedical image segmentations with adversarial loss.
\newblock In \emph{Proceedings of the 27th International Joint Conference on Artificial Intelligence (IJCAI)}, pages 691--697, 2018.

\bibitem[Pizzati et~al.(2020)Pizzati, de~Charette, Zaccaria, and Cerri]{Pizzati2020}
Fabio Pizzati, Raoul de~Charette, Michela Zaccaria, and Pietro Cerri.
\newblock Domain bridge for unpaired image-to-image translation and unsupervised domain adaptation.
\newblock In \emph{WACV}, 2020.

\bibitem[Musto and Zinelli(2020)]{Musto2020}
Luigi Musto and Andrea Zinelli.
\newblock Semantically adaptive image-to-image translation for domain adaptation of semantic segmentation.
\newblock In \emph{BMVC}, 2020.

\bibitem[Xie et~al.(2020)Xie, Chen, Li, Shen, Ma, and Zheng]{Xie2020}
Xinpeng Xie, Jiawei Chen, Yuexiang Li, Linlin Shen, Kai Ma, and Yefeng Zheng.
\newblock Self-supervised cyclegan for object-preserving image-to-image domain adaptation.
\newblock In \emph{ECCV}, 2020.

\bibitem[Gal et~al.(2021)Gal, Patashnik, Maron, Chechik, and Cohen-Or]{gal2021stylegannada}
Rinon Gal, Or~Patashnik, Haggai Maron, Gal Chechik, and Daniel Cohen-Or.
\newblock Stylegan-nada: Clip-guided domain adaptation of image generators, 2021.

\bibitem[Hinton et~al.(2015)Hinton, Vinyals, and Dean]{hinton2015distilling}
Geoffrey Hinton, Oriol Vinyals, and Jeff Dean.
\newblock Distilling the knowledge in a neural network.
\newblock \emph{arXiv:1503.02531}, 2015.

\bibitem[Dhar et~al.(2019)Dhar, Singh, Peng, Wu, and Chellappa]{dhar2019learning}
Prithviraj Dhar, Rajat~Vikram Singh, Kuan-Chuan Peng, Ziyan Wu, and Rama Chellappa.
\newblock Learning without memorizing.
\newblock In \emph{CVPR}, pages 5138--5146, 2019.

\bibitem[Li and Hoiem(2017)]{li2017learning}
Zhizhong Li and Derek Hoiem.
\newblock Learning without forgetting.
\newblock \emph{PAMI}, 40\penalty0 (12):\penalty0 2935--2947, 2017.

\bibitem[Romero et~al.(2014)Romero, Ballas, Kahou, Chassang, Gatta, and Bengio]{romero2014fitnets}
Adriana Romero, Nicolas Ballas, Samira~Ebrahimi Kahou, Antoine Chassang, Carlo Gatta, and Yoshua Bengio.
\newblock Fitnets: Hints for thin deep nets.
\newblock \emph{arXiv:1412.6550}, 2014.

\bibitem[Wang et~al.(2019{\natexlab{b}})Wang, Yuan, Zhang, and Feng]{wang2019distilling}
Tao Wang, Li~Yuan, Xiaopeng Zhang, and Jiashi Feng.
\newblock Distilling object detectors with fine-grained feature imitation.
\newblock In \emph{CVPR}, pages 4933--4942, 2019{\natexlab{b}}.

\bibitem[Zagoruyko and Komodakis(2016)]{zagoruyko2016paying}
Sergey Zagoruyko and Nikos Komodakis.
\newblock Paying more attention to attention: Improving the performance of convolutional neural networks via attention transfer.
\newblock \emph{arXiv:1612.03928}, 2016.

\bibitem[Wang et~al.(2017{\natexlab{b}})Wang, Jiang, Qian, Yang, Li, Zhang, Wang, and Tang]{wang2017residual}
Fei Wang, Mengqing Jiang, Chen Qian, Shuo Yang, Cheng Li, Honggang Zhang, Xiaogang Wang, and Xiaoou Tang.
\newblock Residual attention network for image classification.
\newblock In \emph{CVPR}, pages 3156--3164, 2017{\natexlab{b}}.

\bibitem[Liu et~al.(2020)Liu, Yang, Ravichandran, Bhotika, and Soatto]{liu2020continual}
Xialei Liu, Hao Yang, Avinash Ravichandran, Rahul Bhotika, and Stefano Soatto.
\newblock Continual universal object detection.
\newblock \emph{arXiv:2002.05347}, 2020.

\bibitem[Chen et~al.(2017)Chen, Choi, Yu, Han, and Chandraker]{chen2017learning}
Guobin Chen, Wongun Choi, Xiang Yu, Tony Han, and Manmohan Chandraker.
\newblock Learning efficient object detection models with knowledge distillation.
\newblock In \emph{NIPS}, pages 742--751, 2017.

\bibitem[Gupta et~al.(2016)Gupta, Hoffman, and Malik]{Gupta2016}
Saurabh Gupta, Judy Hoffman, and Jitendra Malik.
\newblock Cross modal distillation for supervision transfer.
\newblock In \emph{CVPR}, 2016.

\bibitem[Zoph et~al.(2020)Zoph, Cubuk, Ghiasi, Lin, Shlens, and Le]{Zoph2020}
Barret Zoph, Ekin~D. Cubuk, Golnaz Ghiasi, Tsung-Yi Lin, Jonathon Shlens, and Quoc~V. Le.
\newblock Learning data augmentation strategies for object detection.
\newblock In \emph{ECCV}, 2020.

\bibitem[Kim et~al.(2021)Kim, Choo, Jeong, and Song]{kim2021comixup}
JangHyun Kim, Wonho Choo, Hosan Jeong, and Hyun~Oh Song.
\newblock Co-mixup: Saliency guided joint mixup with supermodular diversity.
\newblock In \emph{International Conference on Learning Representations}, 2021.

\bibitem[Zhong et~al.(2017)Zhong, Zheng, Kang, Li, and Yang]{zhong2017random}
Zhun Zhong, Liang Zheng, Guoliang Kang, Shaozi Li, and Yi~Yang.
\newblock Random erasing data augmentation.
\newblock \emph{arXiv:1708.04896}, 2017.

\bibitem[Huang et~al.(2020{\natexlab{a}})Huang, Ke, and Huang]{HuangIAN2020}
Zeyi Huang, Wei Ke, and Dong Huang.
\newblock Improving object detection with inverted attention.
\newblock In \emph{WACV}, 2020{\natexlab{a}}.

\bibitem[Huang et~al.(2020{\natexlab{b}})Huang, Wang, Xing, and Huang]{huangRSC2020}
Zeyi Huang, Haohan Wang, Eric~P. Xing, and Dong Huang.
\newblock Self-challenging improves cross-domain generalization.
\newblock In \emph{ECCV}, 2020{\natexlab{b}}.

\bibitem[Chen et~al.(2020)Chen, Tan, Wang, Lu, Hu, and Fu]{chen2020tip}
Shuhan Chen, Xiuli Tan, Ben Wang, Huchuan Lu, Xuelong Hu, and Yun Fu.
\newblock Reverse attention based residual network for salient object detection.
\newblock \emph{TIP}, 29:\penalty0 3763--3776, 2020.

\bibitem[Hou et~al.(2019)Hou, Ma, Liu, and Loy]{hou2019learning}
Yuenan Hou, Zheng Ma, Chunxiao Liu, and Chen~Change Loy.
\newblock Learning lightweight lane detection cnns by self attention distillation.
\newblock In \emph{ICCV}, pages 1013--1021, 2019.

\bibitem[Wang et~al.(2019{\natexlab{c}})Wang, Wu, Karanam, Peng, Singh, Liu, and Metaxas]{wang2019sharpen}
Lezi Wang, Ziyan Wu, Srikrishna Karanam, Kuan-Chuan Peng, Rajat~Vikram Singh, Bo~Liu, and Dimitris~N Metaxas.
\newblock Sharpen focus: Learning with attention separability and consistency.
\newblock In \emph{ICCV}, pages 512--521, 2019{\natexlab{c}}.

\bibitem[Huang et~al.(2020{\natexlab{c}})Huang, Zou, Kumar, and Huang]{huang2020comprehensive}
Zeyi Huang, Yang Zou, BVK Kumar, and Dong Huang.
\newblock Comprehensive attention self-distillation for weakly-supervised object detection.
\newblock \emph{NIPS}, 33, 2020{\natexlab{c}}.

\bibitem[Rabbi et~al.(2020)Rabbi, Ray, Schubert, Chowdhury, and Chao]{rabbi2020small}
Jakaria Rabbi, Nilanjan Ray, Matthias Schubert, Subir Chowdhury, and Dennis Chao.
\newblock Small-object detection in remote sensing images with end-to-end edge-enhanced gan and object detector network.
\newblock \emph{Remote Sensing}, 12\penalty0 (9):\penalty0 1432, 2020.

\bibitem[Liu et~al.(2019{\natexlab{b}})Liu, Muelly, Deng, Pfister, and Li]{LiuMuelly2019}
Lanlan Liu, Michael Muelly, Jia Deng, Tomas Pfister, and Jia Li.
\newblock Generative modeling for small-data object detection.
\newblock In \emph{ICCV}, 2019{\natexlab{b}}.

\bibitem[van~den Oord et~al.(2017)van~den Oord, Vinyals, and Kavukcuoglu]{VQVAE2017}
Aaron van~den Oord, Oriol Vinyals, and Koray Kavukcuoglu.
\newblock Neural discrete representation learning.
\newblock \emph{arXiv:1711.00937}, 2017.

\bibitem[Razavi et~al.(2019)Razavi, van~den Oord, and Vinyals]{VQVAE2019}
Ali Razavi, Aaron van~den Oord, and Oriol Vinyals.
\newblock Generating diverse high-fidelity images with vq-vae-2.
\newblock \emph{arXiv:1906.00446}, 2019.

\bibitem[Esser et~al.(2021)Esser, Rombach, and Ommer]{VQGAN2021}
Patrick Esser, Robin Rombach, and Bjorn Ommer.
\newblock Taming transformers for high-resolution image synthesis.
\newblock \emph{arXiv:2012.09841}, 2021.

\bibitem[Zhou et~al.(2016)Zhou, Khosla, Lapedriza, Oliva, and Torralba]{zhou2016learning}
Bolei Zhou, Aditya Khosla, Agata Lapedriza, Aude Oliva, and Antonio Torralba.
\newblock Learning deep features for discriminative localization.
\newblock In \emph{CVPR}, pages 2921--2929, 2016.

\bibitem[Halperin et~al.(2011)Halperin, Hu, Sheth, and Wetherall]{halperin2011tool}
Daniel Halperin, Wenjun Hu, Anmol Sheth, and David Wetherall.
\newblock Tool release: Gathering 802.11 n traces with channel state information.
\newblock \emph{ACM SIGCOMM Computer Communication Review}, 41\penalty0 (1):\penalty0 53--53, 2011.

\bibitem[Bengio et~al.(1994)Bengio, Simard, and Frasconi]{Bengio1994}
Y.~Bengio, P.~Simard, and P.~Frasconi.
\newblock Learning long-term dependencies with gradient descent is difficult.
\newblock \emph{IEEE Transactions on Neural Networks}, 5\penalty0 (2):\penalty0 157--–166, 1994.

\bibitem[Glorot and Bengio(2010)]{GlorotBengio2010}
X.~Glorot and Y.~Bengio.
\newblock Understanding the difficulty of training deep feedforward neural networks.
\newblock In \emph{AISTATS}, 2010.

\bibitem[Vondrick et~al.(2013)Vondrick, Khosla, Malisiewicz, and Torralba]{vondrick2013hoggles}
C.~Vondrick, A.~Khosla, T.~Malisiewicz, and A.~Torralba.
\newblock {HOGgles: Visualizing Object Detection Features}.
\newblock \emph{ICCV}, 2013.

\bibitem[Cao and Johnson(2021)]{CaoJohnson2021}
Ang Cao and Justin Johnson.
\newblock Inverting and understanding object detectors.
\newblock \emph{arXiv:2106.13933}, 2021.

\bibitem[Park et~al.(2021)Park, Ambrus, Guizilini, Li, and Gaidon]{park2021dd3d}
Dennis Park, Rares Ambrus, Vitor Guizilini, Jie Li, and Adrien Gaidon.
\newblock Is pseudo-lidar needed for monocular 3d object detection?
\newblock In \emph{ICCV}, 2021.

\bibitem[Simonelli et~al.(2020)Simonelli, Bulo, Porzi, Antequera, and Kontschieder]{Simonelli2020}
Andrea Simonelli, Samuel~Rota Bulo, Lorenzo Porzi, Manuel~Lopez Antequera, and Peter Kontschieder.
\newblock Disentangling monocular 3d object detection: From single to multiclass recognition.
\newblock \emph{PAMI}, 2020.

\bibitem[Zhang and Ma(2021)]{ZhangMa2021}
Linfeng Zhang and Kaisheng Ma.
\newblock Improve object detection with feature-based knowledge distillation: Towards accurate and efficient detectors.
\newblock In \emph{ICLR}, 2021.

\end{thebibliography}






\end{document}